\documentclass[11pt]{article}

\PassOptionsToPackage{table}{xcolor}
\usepackage[preprint]{acl}
\usepackage{times}
\usepackage{latexsym}

\usepackage[T1]{fontenc}

\usepackage[utf8]{inputenc}

\usepackage{microtype}

\usepackage{inconsolata}

\usepackage{graphicx}

\usepackage[utf8]{inputenc}
\usepackage{booktabs} 
\usepackage{multirow} 
\usepackage{tabularx} 
\usepackage{amsmath}  
\usepackage{graphicx}
\usepackage{enumerate}
\usepackage{enumitem}
\usepackage{wrapfig}
\usepackage{booktabs}
\usepackage{pifont}
\usepackage{longtable}
\usepackage{caption}
\usepackage{subcaption}
\usepackage{booktabs}
\usepackage{adjustbox}
\usepackage{amssymb} 
\usepackage{float}
\usepackage[table]{xcolor}
\usepackage{arydshln}
\usepackage{colortbl}
\usepackage[most]{tcolorbox}

\newcolumntype{Y}{>{\centering\arraybackslash}X}
\newcommand{\methodname}{CharTide}
\newcommand{\methodnamefull}{Data-Centric Chart-to-Code Generation via \textbf{T}ri-Perspective Tuning and \textbf{I}nquiry-\textbf{D}riven \textbf{E}volution}

\usepackage{xspace}

\title{CharTide: Data-Centric Chart-to-Code Generation via
Tri-Perspective Tuning and Inquiry-Driven Evolution}

\author{
 \textbf{Xiangxi Zheng\textsuperscript{1}},
 \textbf{Kuang He\textsuperscript{2}},
 \textbf{Jiayi Hu\textsuperscript{3}},
 \textbf{Ping Yu\textsuperscript{1}},
 \textbf{Rui Yan\textsuperscript{4}},
 \textbf{Yuan Yao\textsuperscript{1}$^\dagger$},
\\
 \textbf{Peng Hou\textsuperscript{2}},
 \textbf{Anxiang Zeng\textsuperscript{2}},
 \textbf{Alex Jinpeng Wang\textsuperscript{5}$^\dagger$},
\\
\\
 \textsuperscript{1}Nanjing University 
 \textsuperscript{2}LLM Team, Shopee Pte. Ltd.
 \textsuperscript{3}East China Normal University 
\\
 \textsuperscript{4}Nanjing University of Science and Technology 
 \textsuperscript{5}Central South University
\\
 \texttt{zhengxx@smail.nju.edu.cn, y.yao@nju.edu.cn, jinpengwang@csu.edu.cn} 
}

\begin{document}
\maketitle

\renewcommand{\thefootnote}{\fnsymbol{footnote}}
\footnotetext[2]{Corresponding authors: Yuan Yao, Alex Jinpeng Wang}

\begin{abstract}
Chart-to-code generation demands strict visual precision and syntactic correctness from Vision-Language Models (VLMs). 
However, existing approaches are fundamentally constrained by \textbf{data-centric limitations}:
despite the availability of growing chart-to-code datasets, simply scaling homogeneous chart-code pairs conflates visual perception with program logic, preventing models from fully leveraging the richness of multimodal supervision.
We present \methodname{},  a novel \textbf{data-centric framework} that systematically redesigns both training and alignment data for chart-to-code generation.
First, we construct a \textbf{2M-sample} dataset via a \textbf{Tri-Perspective Tuning} strategy, explicitly decoupling training into visual perception, pure-text code logic, and modality fusion streams, enabling a 7B model to surpass specialized baselines using only supervised data.
Second, we reformulate alignment as a \textbf{data verification} problem rather than a heuristic scoring task. To this end, we introduce an \textbf{Inquiry-Driven RL} framework grounded in the principle of information invariance: a downstream model should yield consistent answers to identical visual queries across both original and generated charts.
Moving beyond rigid rule matching or VLM scoring, we employ a frozen Inspector to objectively verify generated charts through atomic QA tasks, providing verifiable reward signals based on answer accuracy.
Experiments on ChartMimic, Plot2Code, and ChartX show that \textbf{\methodname{}-7B/8B} significantly outperforms open-source baselines, surpasses GPT-4o, and is competitive with GPT-5.
\end{abstract}


\section{Introduction}
\label{sec:intro}

\begin{figure}[t]
    \centering
    \includegraphics[width=1.0\columnwidth]{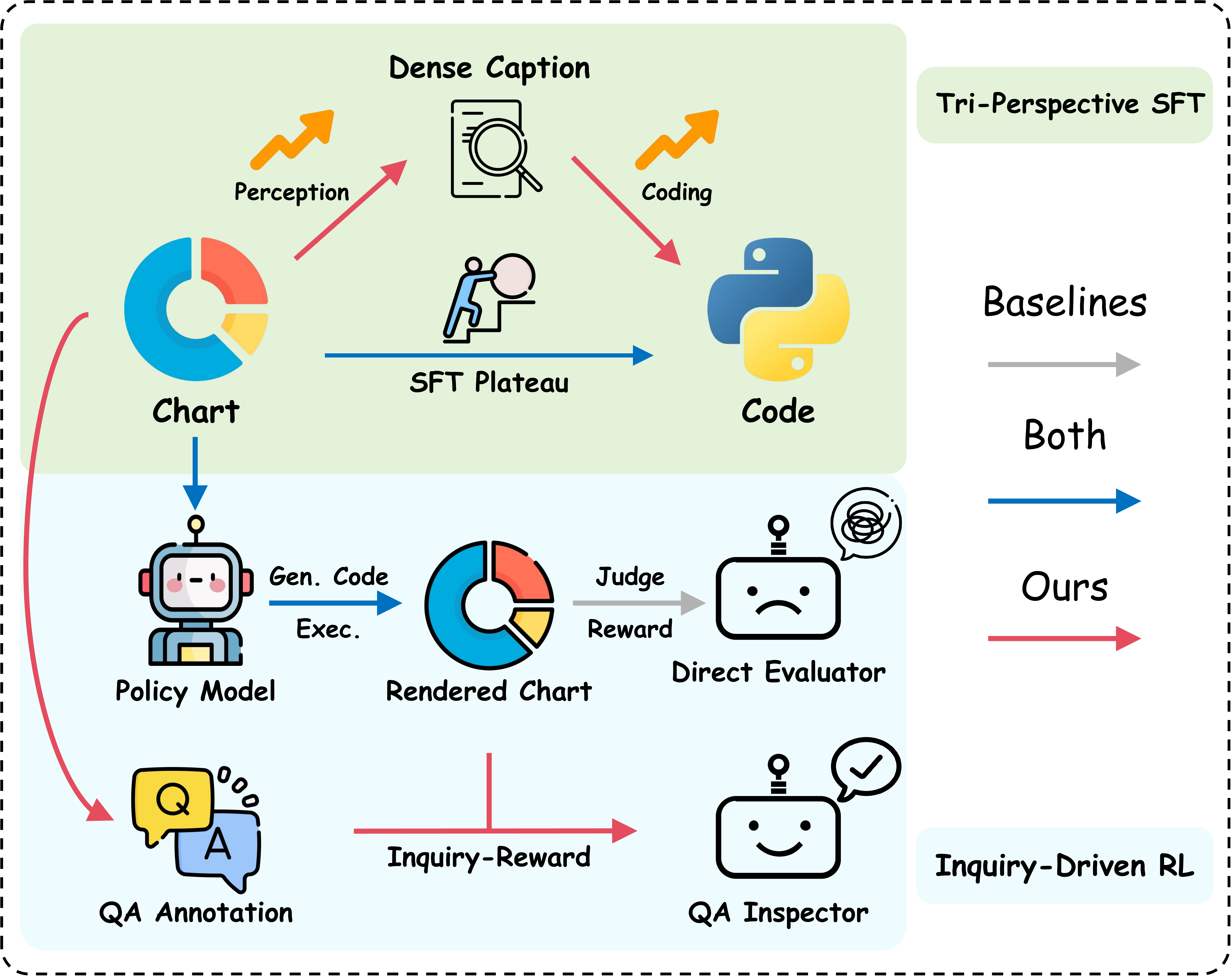}
    \caption{
    \textbf{Overview of the Data-Centric \methodname{} Framework.} 
    (Top) \textbf{Tri-Perspective SFT} explicitly decouples data streams to break the performance plateau. (Bottom) \textbf{Inquiry-Driven RL} replaces subjective scoring with objective, fact-based verification.}
    \label{fig:intro}
\end{figure}

Recent advancements in Multimodal Large Language Models (MLLMs) have revolutionized vision-language understanding~\cite{bai2025qwen3vltechnicalreport, bai2025qwen25vltechnicalreport, wang2025internvl35advancingopensourcemultimodal, masry2022chartqa, masry2025chartqapro}, and have further extended to multimodal code generation for UIs and sketches~\cite{si2025design2codebenchmarkingmultimodalcode, jiang2025screencoderadvancingvisualtocodegeneration, Wan_2025, yun2024web2codelargescalewebpagetocodedataset, gui2025webcode2m}. 
However, applying such capabilities to \textbf{Chart-to-Code generation} presents a unique challenge stemming from the fundamental difference in data characteristics.
Unlike open-ended chart understanding~\cite{kondic2025chartgenscalingchartunderstanding, he2025distillvisualchartreasoning, xing2025caprlstimulatingdenseimage, chen2025chartr1}, chart-to-code data requires modeling the tight coupling between continuous numerical values and precise visual styling.
This task requires reverse-engineering these dense visual signals into rigorous plotting code under zero-tolerance constraints on both visual precision and syntactic correctness.
Consequently, standard data paradigms effective in general domains fall short here, as they demand that models simultaneously possess fine-grained perception and precise code synthesis capabilities~\cite{yang2025chartmimicevaluatinglmmscrossmodal, xia2025chartxchartvlmversatile} to handle such strict information mapping.



While recent efforts have shifted to real-world data, the Supervised Fine-Tuning (SFT) paradigm has hit a distinct \textbf{scaling wall}. 
Empirical studies such as MSRL~\cite{chen2025breakingsftplateaumultimodal}, which scales to 3M samples, indicate that merely accumulating homogeneous data (i.e., Chart-Code pairs) yields diminishing returns.
We attribute this bottleneck not only to data volume but also to the \textit{inherent inefficiency} of the singular Chart-Code pair format.
Fundamentally, plotting code is a highly structured modality in which boilerplate syntax and non-visual logic occupy a disproportionate share of the token space, thereby diluting supervision for key visual attributes.
This \textbf{information asymmetry} biases models towards \textbf{template memorization} during end-to-end training, leading them to prioritize syntactic patterns rather than achieving deep alignment with fine-grained visual perception.

A similar insufficiency in supervision signals extends to the Reinforcement Learning (RL) stage.
Due to the absence of precise and verifiable evaluation mechanisms, existing approaches rely on subjective, black-box VLM scoring ~\cite{chen2025breakingsftplateaumultimodal} or rigid rule-based matching~\cite{tan2025chartmasteradvancingcharttocodegeneration}.
Consequently, they fail to \textbf{leverage} the \textbf{diverse high-level semantics} embedded in high-quality data, resulting in supervision that is unstable, difficult to reproduce, and data-inefficient.

To address these challenges, we propose \methodnamefull{} (\textbf{\methodname{}}). 
First, we introduce a \textbf{Tri-Perspective Decomposed SFT} strategy to overcome the scaling wall. 
Recognizing that merely scaling data fails to disentangle visual and logical hallucinations~\cite{chen2025breakingsftplateaumultimodal}, we construct orthogonal data streams that decouple training into fine-grained perception, pure-text code logic, and modality fusion. 
This decomposition effectively breaks the homogeneity bottleneck, enabling our 7B-scale model to distill capabilities that surpass 235B-scale baselines on ChartMimic~\cite{yang2025chartmimicevaluatinglmmscrossmodal}.

Furthermore, we propose \textbf{Chart2Code Inquiry-Driven RL}, shifting the alignment paradigm from subjective preference to verifiable fact-checking. 
Grounded in the hypothesis that accurate reproduction ensures the fidelity of information transfer, we replace unstable black-box VLM scoring~\cite{chen2025breakingsftplateaumultimodal, zhang2025boostingcharttocodegenerationmllm} with a deterministic Inspector. 
This mechanism objectively verifies generated charts via rigorous QA pairs and visual constraints, ensuring the generated code is not just semantically precise in data trends but also maintains high pixel-level fidelity.

Experiments on three authoritative benchmarks, including ChartMimic~\cite{yang2025chartmimicevaluatinglmmscrossmodal}, Plot2Code~\cite{wu2024plot2codecomprehensivebenchmarkevaluating}, and ChartX~\cite{xia2025chartxchartvlmversatile}, demonstrate that \methodname{} achieves state-of-the-art (SOTA) performance among open-source models, surpassing GPT-4o and achieving performance comparable to GPT-5.

Our main contributions are as follows:
\begin{itemize}[left=0.2em]
\setlength{\itemsep}{0.05em}
\setlength{\parskip}{0.1em}
\item We propose Tri-Perspective Decomposed SFT to address data homogeneity by decoupling visual perception, code logic, and modality fusion. A curated 2M-sample dataset enables 7B models to acquire complementary chart-to-code capabilities and outperform 235B-scale baselines.

\item We integrate Inquiry-Driven verifiable rewards by reformulating alignment as data verification. By leveraging a frozen Inspector to objectively verify outputs via curated QA data, our approach substantially improves semantic consistency and pixel-level fidelity of the generated code.

\item We present \textbf{\methodname{}}, achieving SOTA performance among open-source models and demonstrating competitive performance with GPT-5. Extensive experiments validate the efficacy of our two-stage pipeline, offering robust insights for future advancements in this domain.

\end{itemize}

\vspace{-5pt}

\begin{figure*}[ht]
    \centering
    \includegraphics[width=1.0\textwidth]{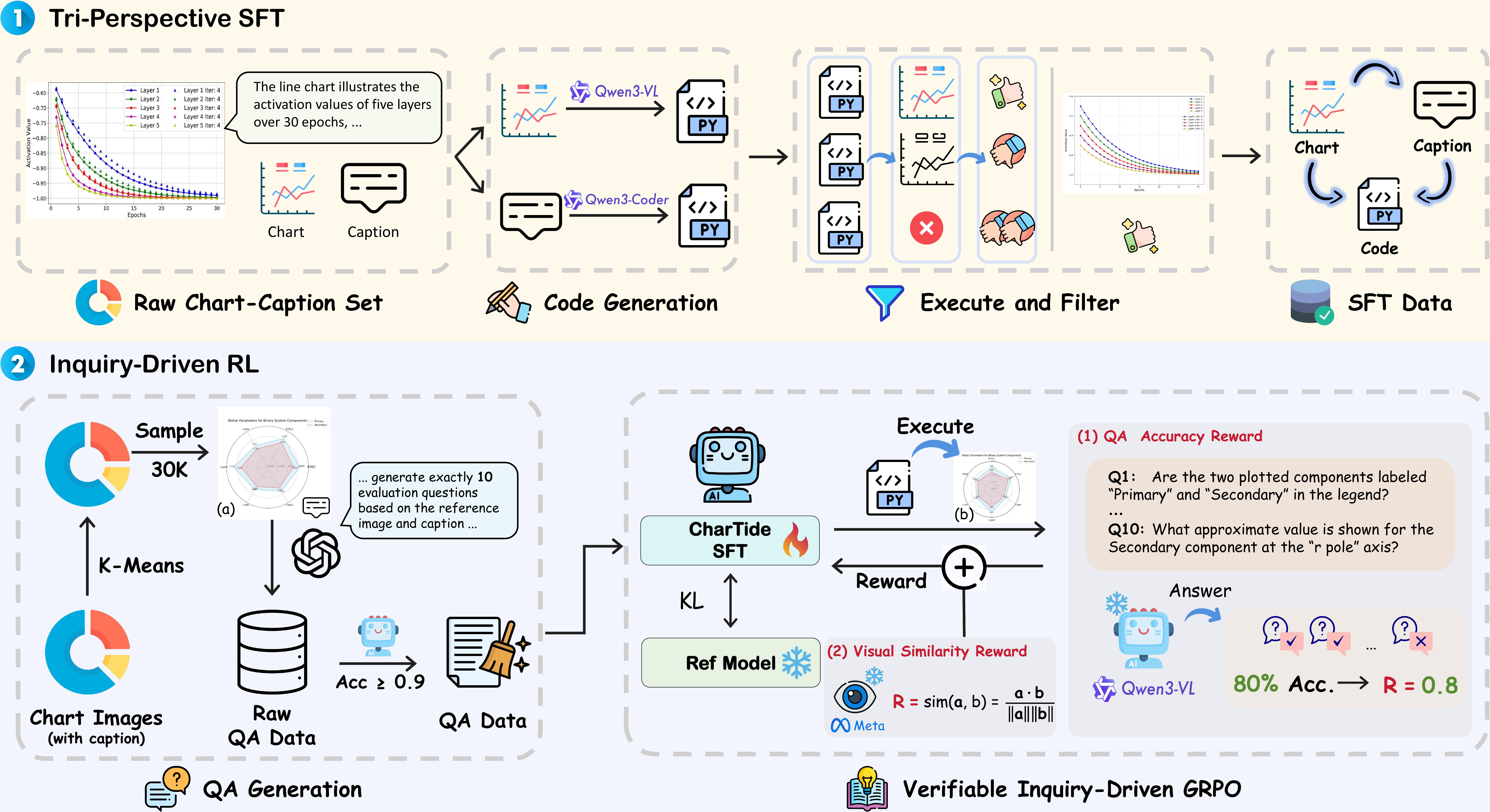}
    \caption{\textbf{The detailed pipeline of \methodname{}.} The pipeline consists of two stages: (1) \textbf{Tri-Perspective SFT} constructs three complementary data streams, including Visual Perception, Code Logic, and Modality Fusion to distill multi-dimensional capabilities into the foundational model; (2) \textbf{Inquiry-Driven RL} aligns the model using a hybrid verification loop, where a frozen Inspector provides objective semantic rewards ($r_{QA}$) via fact-checking, complemented by visual similarity rewards ($r_{vis}$) to ensure rigorous pixel-level and semantic alignment.}
    
\label{fig:pipeline}
    \label{fig:overview}
\end{figure*}
\section{Related Work}
\label{sec:related}

\subsection{Chart-to-Code Generation}

The Chart-to-Code task demands pixel-level visual reproduction through executable code, imposing dual constraints on visual perception and syntactic correctness~\cite{yang2025chartmimicevaluatinglmmscrossmodal, wu2024plot2codecomprehensivebenchmarkevaluating, xia2025chartxchartvlmversatile}. 
Representative chart-to-code methods~\cite{han2023chartllamamultimodalllmchart,zhao2025chartcoderadvancingmultimodallarge} mainly formulate the task as end-to-end supervised generation from charts to plotting code, typically using synthetic or automatically constructed chart--code pairs.
However, synthetic datasets inherently suffer from limited stylistic diversity. While recent works~\cite{tan2025chartmasteradvancingcharttocodegeneration,zhao2025vincicoderunifyingmultimodalcode, chen2025breakingsftplateaumultimodal, niu2025chart2code53, jiang2025viscodex} have pivoted to real-world annotations to improve diversity, concurrent research~\cite{chen2025breakingsftplateaumultimodal} identifies a scaling wall in the SFT paradigm: mere data accumulation without capability decoupling fails to address the entanglement of visual and logical hallucinations.

To overcome the one-to-many mapping ambiguity in code generation, RL has been widely adopted~\cite{zhang2025boostingcharttocodegenerationmllm, tan2025chartmasteradvancingcharttocodegeneration, chen2025breakingsftplateaumultimodal}. 
Existing RL strategies fall into two categories, yet both face significant limitations: 

\noindent(1) \textbf{Rule-based methods}~\cite{tan2025chartmasteradvancingcharttocodegeneration, zhang2025boostingcharttocodegenerationmllm} rely on heuristic attribute matching (e.g., colors, legends), which neglects holistic visual semantics and generalizability; 

\noindent(2) \textbf{VLM-as-a-Judge methods}~\cite{chen2025breakingsftplateaumultimodal} utilize visual similarity scores from large VLMs. 
However, these approaches struggle with the subjective, black-box nature of VLM scoring, leading to high variance and computational costs.
In contrast, \methodname{} introduces an objective, verification-based reward mechanism to resolve these issues.

\subsection{Reinforcement Learning for MLLMs}
\label{sec:rel_rl}

The success of reasoning models such as DeepSeek-R1 ~\cite{deepseekai2025deepseekr1incentivizingreasoningcapability} has ignited significant interest within the multimodal domain regarding Reinforcement Learning with Verifiable Rewards (RLVR)~\cite{zhang2025survey, lambert2025tulu3pushingfrontiers, chen2025learningimagesvisualreinforcement, zha2025vision, chen2025chartr1}. 
Unlike traditional alignment strategies based on human preferences, RLVR optimizes policies leveraging objective and deterministic outcomes. 
Pioneering works ~\cite{huang2025visionr1incentivizingreasoningcapability, meng2025mmeurekaexploringfrontiersmultimodal, yang2025r1onevision} utilize GRPO combined with accuracy-based rewards, successfully fostering emergent capabilities in multimodal reasoning. 
In other adaptations, CapRL~\cite{xing2025caprlstimulatingdenseimage} pioneered the use of downstream task QA performance to implement verifiable rewards for image captioning models, surpassing simple text matching metrics.
Inspired by this, researchers have explored various reward forms for different tasks~\cite{xing2025caprlstimulatingdenseimage, ni2025point}, offering new directions for future research.

\section{Method}
\label{sec:method}

We present \textbf{\methodname{}}, a data-centric framework for precise Chart-to-Code generation. 
As illustrated in Figure~\ref{fig:pipeline}, the framework initially establishes a robust foundation via \textbf{Tri-Perspective Decomposed SFT}, followed by semantic and pixel-level data verification through \textbf{Inquiry-Driven RL}.

\subsection{Tri-Perspective Decomposed SFT}

To overcome the bottlenecks caused by data homogeneity, we leverage charts and high-quality captions from ChartCap~\cite{lim2025chartcapmitigatinghallucinationdense} as our core data source. 
Via multi-view distillation, we construct three \textbf{complementary} training streams, each targeting a distinct capability dimension:

\noindent{\textbf{Visual Perception Stream (Chart $\to$ Caption).}} 
To bridge the perception gap inherent in 7B-scale models, we focus on strictly aligning visual features with dense textual descriptions from ChartCap. 
We apply a simple length-based filter to exclude excessively long captions, ensuring the model focuses on concise and effective visual grounding without context overflow.

\noindent{\textbf{Code Logic Stream (Caption $\to$ Code).}} 
To decouple syntax learning from visual perception, we construct a pure-text code generation stream. 
We prompt Qwen3-Coder-30B-A3B~\cite{yang2025qwen3technicalreport} with detailed captions to generate plotting code that strictly reflects the textual descriptions. 
To ensure quality, we execute and filter the outputs through a visual consistency check using Qwen3-VL-235B-A22B~\cite{bai2025qwen3vltechnicalreport} given the caption and original chart. 
This ensures that we retain only high-quality samples to strengthen the model's logic proficiency without visual interference.

\noindent{\textbf{Modality Fusion Stream (Chart $\to$ Code).}} 
To establish end-to-end generation capability, we integrate 500k samples from ChartCap with publicly available chart datasets~\cite{zhao2025chartcoderadvancingmultimodallarge, zhao2025vincicoderunifyingmultimodalcode}, totaling 1M chart images. 
We utilize Qwen3-VL-235B-A22B to generate plotting code from the source charts, filtering the outputs based on WebSSL feature similarity. 
Instead of pairing the code with the original source images, we strictly pair the generated code with its corresponding rendered image. 
This strategy eliminates potential discrepancies between the visual chart and the code, ensuring strict pixel-level correspondence and mitigating noise from imperfect ground-truth code.

Ultimately, we combine these three streams with open-source instruction data to construct $\mathcal{D}_{SFT}$ ($\approx$ 2M samples), performing full-parameter fine-tuning on Qwen2.5-VL-7B-Instruct and Qwen3-VL-8B-Instruct. The detailed data construction pipeline is provided in Appendix~\ref{sec:appendix_data}.

\begin{table*}[ht]
\centering
\small
\caption{
Comparison with Closed-Source and Open-Source Models on ChartMimic, Plot2Code, and ChartX.  
*For Plot2Code, we change the evaluator from GPT-4V to GPT-4o to ensure robust and reproducible scoring. 
We report normalized scores over the full test set to avoid survivorship bias. 
Details are provided in \textbf{Appendix ~\ref{sec:appendix_eval}}.
}
\label{tab:main_results}
\renewcommand{\arraystretch}{1.15}
\setlength{\tabcolsep}{4pt} 

\resizebox{\textwidth}{!}{
\begin{tabular}{l ccc ccc c}
\toprule
\multirow{2}{*}{\textbf{Model}} & \multicolumn{3}{c}{\textbf{ChartMimic}} & \multicolumn{3}{c}{\textbf{Plot2Code*}} & \textbf{ChartX} \\
\cmidrule(lr){2-4} \cmidrule(lr){5-7} \cmidrule(lr){8-8}
 & Exec Rate & Low Level & High Level & Exec Rate & Text Match & Rating & GPT score \\
\midrule
\textit{Full Score} & \textit{100} & \textit{100} & \textit{100} & \textit{100} & \textit{100} & \textit{10} & \textit{5} \\
\midrule


\multicolumn{8}{c}{\textit{\textbf{Proprietary Models}}} \\
\midrule
\rowcolor{gray!20} GPT-4o & 94.7& 80.0 & 87.7 & 87.1 & 52.6 & 5.66 & 2.61 \\
\rowcolor{gray!20} GPT-5 & 96.8 & 82.1 & 94.7 & 87.8 & 61.9 & 7.28 &  3.59 \\
\rowcolor{gray!20} Gemini-2.5-Pro & 94.7 & 79.2 & 92.5 & 88.6 & 69.1 & 7.45 & 3.27 \\
\midrule

\multicolumn{8}{c}{\textit{\textbf{Open-Source General-Domain}}} \\
\midrule
Qwen2.5-VL-7B & 75.0 & 49.0 & 51.8 & 68.9 & 33.7 & 3.04 & 2.74 \\
Qwen2.5-VL-72B & 75.3 & 51.9 & 56.6 & 59.3 & 33.2 & 3.61 & 2.85 \\
Qwen3-VL-8B & 81.7 & 63.7 & 71.5 & 76.5 & 36.3 & 3.91 & 2.93 \\
Qwen3-VL-30B-A3B & 85.2 & 67.7 & 76.5 & 87.1 & 46.7 & 4.85 & 2.73 \\
Qwen3-VL-235B-A22B & 93.3 & 76.8 & 87.6 & 84.8 & 46.2 & 5.19 & \textbf{3.35} \\
\midrule

\multicolumn{8}{c}{\textit{\textbf{Open-Source Chart-Domain}}} \\
\midrule
ChartCoder-7B & 89.5 & 72.1 & 78.5 & 68.9 & 31.1 & 2.73 & 2.79 \\ 
ChartMaster-7B & 93.5 & 77.1 & 83.3 & \underline{89.4} & 53.6 & 4.73 & 2.82 \\
MSRL-7B-SFT & 92.6 & 71.2 & 82.8 & 77.3 & 34.7 & 3.71 & 3.19 \\
MSRL-7B & 94.3 & 76.1 & 87.4 & 62.9 & 31.9 & 3.24 & 3.22 \\
VinciCoder-7B & 91.2 & 77.0 & 83.4 & 68.9 & 33.6 & 3.39 & 3.18 \\ 
VinciCoder-8B & 90.2 & 75.8 & 81.4 & 85.6 & 49.8 & 4.49 & 3.21 \\
\midrule
\rowcolor{yellow!12} \textbf{\methodname{}-7B-SFT} & 94.3 & 79.3 & 86.4 & 88.6 & 58.2 & 5.17 & 3.00 \\
\rowcolor{yellow!12} \textbf{\methodname{}-7B} & \underline{96.7} & \underline{81.7} & \underline{91.6} & \underline{89.4} & \underline{59.6} & \underline{5.60} & 3.22 \\
\rowcolor{yellow!12} \textbf{\methodname{}-8B-SFT} & 93.7 & 80.9 & 89.4 & 86.4 & 58.1 & 5.46 & 3.19 \\
\rowcolor{yellow!12} \textbf{\methodname{}-8B} & \textbf{97.3} & \textbf{83.0} & \textbf{92.7} & \textbf{91.7} & \textbf{64.6} & \textbf{5.93} & \underline{3.23} \\
\bottomrule
\end{tabular}
}
\end{table*}

\subsection{Inquiry-Driven RL}
\label{sec:rl_method}

Although SFT endows the model with foundational capabilities, it struggles to handle long-tail details. The teacher-forcing paradigm restricts the model's exploration capabilities and output diversity within the code space. Traditional RL approaches face a dilemma: rule-based matching relies on rigid templates, while VLM-as-a-Judge methods suffer from subjective black-box noise. To provide reliable and verifiable supervision, we propose a hybrid verification loop, as illustrated in Figure~\ref{fig:overview}.

\paragraph{VQA Data Preparation.}
To provide high-quality supervisory signals, we constructed a specialized Chart-VQA dataset. 
First, we performed K-Means clustering on 500k ChartCap samples using WebSSL-1B features to select 30k representative charts. We then employed GPT-5 to generate a diverse set of QA pairs ($N=10$ per image), covering dimensions such as titles, legends, and numerical trends, with explicit tolerance annotations for numerical values. 
To mitigate the inherent hallucination noise of the Inspector, we implemented consistency filtering prior to RL: we utilized Qwen3-VL-30B-A3B to pre-screen the VQA data, retaining only chart samples where the model successfully answers at least 9 questions correctly (Acc $\ge$ 0.9). 
This ensures the objectivity of the reward signal by validating the Inspector's understanding of the ground truth. 
Ultimately, we retained approximately 20k images and high-quality QA data for reinforcement learning, denoted as $\mathcal{D}_{RL} = \{ (I_{src}, \mathcal{Q}) \}$, where $\mathcal{Q} = \{(q_i, a_i)\}_{i=1}^N$. Further details regarding the data construction pipeline are provided in Appendix~\ref{sec:appendix_qa_gen}.

\paragraph{Inquiry-Driven Rewards.}
During training, we sample $(I_{src}, \mathcal{Q}) \sim \mathcal{D}_{RL}$. The policy $\pi_\theta$ generates code to render a predicted image $I_{pred}$. We define the semantic consistency reward $r_{QA}$ as the pass rate verified by the Inspector:
\begin{equation}
    r_{QA} = \frac{1}{|\mathcal{Q}|} \sum_{(q, a) \in \mathcal{Q}} \mathbb{I}\left( \mathcal{M}(\text{Inspector}(I_{pred}, q), a) \right)
\end{equation}
where $\mathbb{I}(\cdot)$ is the indicator function and $\mathcal{M}$ checks semantic alignment incorporating numerical tolerance. While $r_{QA}$ ensures semantic accuracy, the ``One-to-Many'' nature of code generation may lead to visually suboptimal styling. To impose structural constraints, we introduce a visual consistency reward $r_{vis}$ based on WebSSL-1B, which we empirically found superior to DINO or SigLIP in detecting structural collapse:
\begin{equation}
    r_{vis} = \text{CosineSim}(\text{Enc}_{web}(I_{src}), \text{Enc}_{web}(I_{pred}))
\end{equation}

The total reward is computed as $R_{total} = r_{QA} + \lambda \cdot r_{vis}$, where $\lambda$ is a balancing coefficient. 
We optimize the policy $\pi_\theta$ using Group Relative Policy Optimization (GRPO)~\cite{shao2024deepseekmathpushinglimitsmathematical}. 
For each query, we sample a group of outputs $\{o_i\}_{i=1}^G$ and compute the advantage $\hat{A}_i$ by standardizing the total rewards within the group. 
The model is then updated to maximize this group-relative advantage, incorporating a KL divergence penalty to constrain the policy updates within a stable trust region.

In contrast to VLM-as-a-Judge paradigms that suffer from subjective holistic scoring, our approach simplifies evaluation into atomic verification tasks.
By decoupling the evaluator's perception from the generator's performance via ground-truth pre-filtering, we effectively transform stochastic black-box assessments into deterministic, low-variance supervision signals.

\section{Experiments}
\label{sec:experiments}

\subsection{Implementation Details}
We build \methodname{}-7B and 8B on Qwen2.5-VL-7B and Qwen3-VL-8B backbones, respectively. \\
\textbf{SFT Stage:} 
We perform full-parameter fine-tuning on the 2M multi-perspective dataset. 
We set the global batch size to 256, the initial learning rate to $1\mathrm{e}{-5}$. 
This process requires approximately 36 hours on 8 $\times$ H100 GPUs. \\
\textbf{RL Stage:} 
We initialize the policy with the SFT checkpoint and optimize it using the \textbf{$\approx$ 20k} verified VQA samples. 
We set the global batch size to 128, the learning rate to $1\mathrm{e}{-6}$, and the KL penalty coefficient $\beta=0.02$. 
Training is conducted on 8 $\times$ H100 GPUs for the policy model, with an additional 4 $\times$ H100 GPUs allocated for the inference of the frozen Inspector and WebSSL reward models. 
The RL training completes in roughly 20 hours.

\subsection{Main Results}
\label{sec:main_results}
We evaluate \methodname{} against state-of-the-art proprietary and open-source models on ChartMimic~\cite{yang2025chartmimicevaluatinglmmscrossmodal}, Plot2Code~\cite{wu2024plot2codecomprehensivebenchmarkevaluating}, and ChartX~\cite{xia2025chartxchartvlmversatile}.
As shown in Table~\ref{tab:main_results}, \methodname{} achieves SOTA performance, surpassing both large general VLMs and specialized chart-to-code models, including MSRL~\cite{chen2025breakingsftplateaumultimodal} and ChartMaster~\cite{tan2025chartmasteradvancingcharttocodegeneration}.
On ChartMimic, \methodname{}-7B attains a High-Level score of \textbf{91.6}, outperforming the previous best open-source model MSRL-7B (87.4) and \textbf{GPT-4o} (87.7).
This trend holds on Plot2Code and ChartX, where \methodname{} consistently leads open-source models and remains competitive with top-tier proprietary systems such as \textbf{GPT-5}.
Overall, these results demonstrate that \methodname{} effectively bridges the capability gap between open-weights and top-tier closed-source models.

\subsection{Ablation Study}

\subsubsection{SFT and RL Ablation}
We conduct comprehensive ablation studies on ChartMimic to evaluate the contribution of each component within \methodname{}.

\paragraph{SFT Strategy Ablation.}
We evaluate the effectiveness of the Tri-Perspective decoupled data strategy in Table~\ref{tab:sft_ablation}. Rows 1-3 show that scaling homogeneous Chart-to-Code (C2C) data quickly saturates, with marginal gains when increasing data from 800K to 1M.
To overcome this bottleneck, we progressively introduce decoupled data streams.
Adding 500k Image-Caption data (Row 4) improves Low/High-level scores while preserving execution rate, evidencing an enhancement in fine-grained visual perception. 
Further incorporating 400k Caption-to-Code data (Row 5) markedly increases the \textbf{Execution Rate (92.5 $\rightarrow$ 94.3)}, confirming that pure-text logic training effectively isolates syntax learning from visual noise.
Overall, the complementary streams jointly contribute to performance gains across multiple dimensions.
\begin{table}[t]
    \centering
    \caption{\textbf{Ablation of SFT Data Strategies.} We verify the impact of incrementally adding data sources. \textbf{C2C}: Chart-to-Code; \textbf{Cap}: Chart-to-Caption ; \textbf{Cap2C}: Caption-to-Code. Incremental integration of decoupled training streams yields consistent performance gains.} 
    \label{tab:sft_ablation}
    \resizebox{\columnwidth}{!}{
        \setlength{\tabcolsep}{10pt}
        \renewcommand{\arraystretch}{1.2}
        \begin{tabular}{ccc ccc}
            \toprule
            \multicolumn{3}{c}{\textbf{Data Source}} & \multicolumn{3}{c}{\textbf{ChartMimic}} \\
            \cmidrule(lr){1-3} \cmidrule(lr){4-6}
            \textbf{C2C} & \textbf{Cap} & \textbf{Cap2C} & \textbf{Exec R.} & \textbf{Low L.} & \textbf{High L.} \\
            \midrule
            400K & - & - & 84.5 & 68.9 & 75.7 \\
            800K & - & - & 91.3 & 77.5 & 85.3 \\
            1M & - & - & \cellcolor{blue!8}92.0 & \cellcolor{blue!8}77.6 & \cellcolor{blue!8}85.1 \\
            1M & 500K & - & \cellcolor{blue!15}92.5 &  \cellcolor{blue!15}78.8 & \cellcolor{blue!15}86.4 \\
            1M & 500K & 400K & \cellcolor{blue!22}\textbf{94.3} & \cellcolor{blue!22}\textbf{79.3} & \cellcolor{blue!22}\textbf{87.4} \\
            \bottomrule
        \end{tabular}
    }
    \vspace{-10pt}
\end{table}

\paragraph{Impact of SFT Basis and RL Universality.}
Table~\ref{tab:sft_rl_impact} illustrates the symbiosis between SFT and RL. We observe that directly applying RL to the raw Qwen2.5-VL base model (Row 4) improves code execution but \textbf{fails to achieve holistic alignment}, lagging significantly behind the SFT baseline in visual fidelity (76.7 vs 86.4 High-Level). 
This confirms that \textbf{a robust SFT foundation} is essential for initializing the policy space to a learnable region. 
However, once properly initialized via SFT, our Inquiry-Driven RL substantially boosts performance, pushing the High-Level score to 91.6. 
Furthermore, this strategy proves effective even when applied to external baselines. Specifically, applying our RL stage to the \textbf{ChartMaster-7B}~\cite{tan2025chartmasteradvancingcharttocodegeneration} checkpoint (Row 2) yields further performance gains. 
This demonstrates that our Inquiry-Driven reward is a versatile alignment framework capable of refining diverse foundations, regardless of their underlying training recipes.

\begin{table}[t]
    \centering
    \caption{\textbf{Impact of SFT Basis and RL Universality.} We validate the effectiveness of RL on different foundations, including third-party models (ChartMaster) and different training stages.}
    \label{tab:sft_rl_impact}
    \resizebox{\columnwidth}{!}{
        \setlength{\tabcolsep}{8pt}
        \renewcommand{\arraystretch}{1.2}
        \begin{tabular}{l cc ccc}
            \toprule
            \multirow{2}{*}{\textbf{Base Model}} & \multicolumn{2}{c}{\textbf{Stage}} & \multicolumn{3}{c}{\textbf{ChartMimic}} \\
            \cmidrule(lr){2-3} \cmidrule(lr){4-6}
             & \textbf{SFT} & \textbf{RL} & \textbf{Exec R.} & \textbf{Low L.} & \textbf{High L.} \\
            \midrule
            \multirow{2}{*}{ChartMaster-7B} & & & 93.5 & 77.1 & 83.3 \\
             & & \checkmark  & \cellcolor{blue!6}{96.0} & \cellcolor{blue!6}{79.2} & \cellcolor{blue!6}{84.6} \\
            \midrule
            \multirow{4}{*}{Qwen2.5-VL-7B} &  &  & 75.0 & 49.0 & 51.8 \\
             &  & \checkmark & 94.5 & 68.0 & 76.7  \\
             & \checkmark &  & \cellcolor{blue!8}94.3 & \cellcolor{blue!8}79.3 & \cellcolor{blue!8}86.4 \\
             \rowcolor{blue!15}\cellcolor{white} & \cellcolor{white}\checkmark & \cellcolor{white}\checkmark & \textbf{96.7} & \textbf{81.7} & \textbf{91.6} \\
            \bottomrule
        \end{tabular}
    }
\end{table}

\begin{table}[t]
    \vspace{-2pt}

    \centering
    \caption{\textbf{Ablation of Visual Encoders and Reward Mechanisms.} Section (a) compares visual backbones using only $R_{vis}$. Section (b) fixes the backbone to WebSSL-1B and the inspector to Qwen3-VL-30B-A3B to verify the effectiveness of our Inquiry-Driven strategy against VLM-Judge baselines.}
    \label{tab:ablation_simplified}
    
    \resizebox{\columnwidth}{!}{
        \setlength{\tabcolsep}{6pt}
        \renewcommand{\arraystretch}{1.2}
        \begin{tabular}{ll ccc}
            \toprule
            \multirow{2}{*}{\textbf{Settings}} & \textbf{Backbone / } & \multicolumn{3}{c}{\textbf{ChartMimic}} \\
            \cmidrule(lr){3-5}
             & \textbf{Inspector} & \textbf{Exec R.} & \textbf{Low L.} & \textbf{High L.} \\
            \midrule
            \multicolumn{5}{l}{\textit{(a) Visual Encoder Ablation (using $R_{vis}$ only)}} \\
            \midrule
            \multirow{3}{*}{$R_{vis}$} & SigLIP-so400m & 95.5 & 79.2 & 88.7 \\
            & DINOv2-G & 96.0 & 80.2 & 89.2 \\ 
            & \textbf{WebSSL-1B} & \cellcolor{blue!8}96.4 & \cellcolor{blue!8}80.5 & \cellcolor{blue!8}89.8 \\
            \midrule
            \multicolumn{5}{l}{\textit{(b) Reward Mechanism Ablation ($R_{vis}$ with WebSSL-1B)}} \\
            \midrule
            $R_{judge}$ & \multirow{4}{*}{\shortstack[l]{Q3-VL-30B}} & 95.5 & 79.9 & 89.0 \\
            $R_{inq}$ &  & \cellcolor{blue!8}96.5 & \cellcolor{blue!8}80.4 & \cellcolor{blue!8}89.3 \\
            \cmidrule(lr){1-1} \cmidrule(lr){3-5}
            $R_{vis} + R_{judge}$ &  & 96.0 & 80.3 & 90.9 \\
            $R_{vis} + R_{inq}$ (\textbf{Ours}) &  & \cellcolor{blue!15}\textbf{96.7} & \cellcolor{blue!15}\textbf{81.7} & \cellcolor{blue!15}\textbf{91.6} \\
            \bottomrule
        \end{tabular}
    }
    \vspace{-7pt}
    
\end{table}

\begin{figure*}[t]
    \centering
    \includegraphics[width=1.0\textwidth]{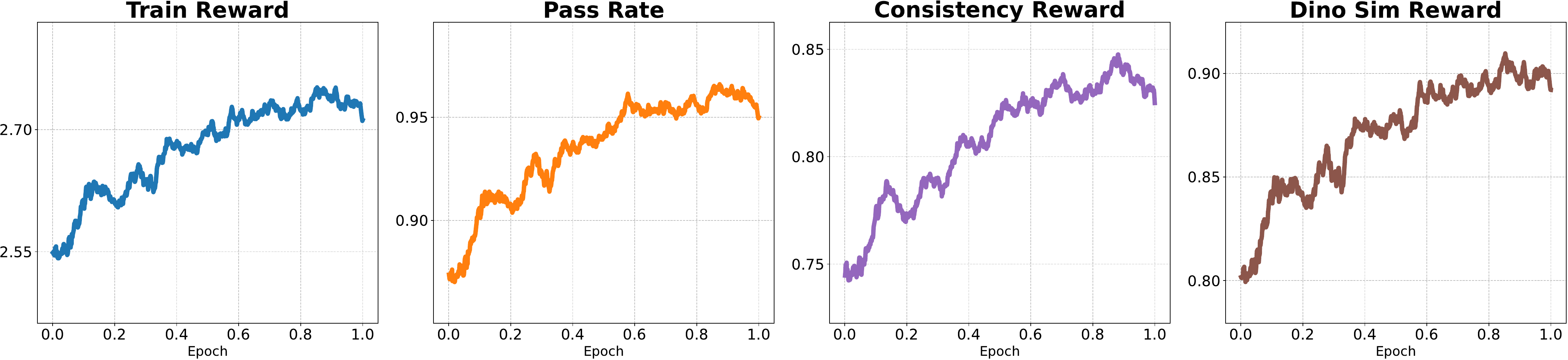}
    \small
    \caption{\textbf{Training dynamics of the Inquiry-Driven RL phase.} The Train Reward, Pass Rate, and Consistency Reward exhibit synchronized upward trajectories, indicating stable and effective optimization.}
    \label{fig:train_curves}
    \vspace{-10pt}
\end{figure*}

\subsubsection{Reward Ablation}
We conduct an in-depth ablation analysis of the core reward mechanisms to validate our design choices. The results are summarized in Table~\ref{tab:ablation_simplified}.

\vspace{-3pt}
\paragraph{Visual Encoder Selection}
Section (a) evaluates the impact of different visual backbones when using a pure visual similarity reward ($R_{vis}$). 
We compare the semantic-focused \textbf{SigLIP-so400m}~\cite{zhai2023sigmoid} against two structure-aware SSL models: \textbf{DINOv2-G}~\cite{oquab2023dinov2} and \textbf{WebSSL-1B}~\cite{fan2025scalinglanguagefreevisualrepresentation}.
We observe that WebSSL-1B achieves the highest Execution Rate (\textbf{96.4}) and Low-Level score (\textbf{80.5}).
Although it shares the same DINO-based architecture as DINOv2, WebSSL derives significant benefits from being pre-trained on large-scale web data. 
This domain alignment makes it more effective than natural-image-based models in capturing the fine-grained structural nuances (e.g., grid lines, marker shapes) required for precise chart reproduction. We provide more qualitative comparisons in Appendix~\ref{sec:appendix_webssl}.

\vspace{-3pt}
\paragraph{Reward Strategies Comparison}
Section (b) validates the effectiveness of our Inquiry-Driven strategy ($R_{inq}$) compared to the VLM Judge baseline ($R_{judge}$). The comparison yields two key insights:

\noindent (1) \textbf{\textit{Instability of VLM-as-a-Judge.}} Relying solely on the VLM judge ($R_{judge}$) yields the lowest performance across most metrics. Moreover, adding $R_{judge}$ to $R_{vis}$ causes regression in Low-Level metrics ($80.5 \to 80.3$) and Execution Rate ($96.4 \to 96.0$).
This confirms that subjective, black-box VLM scoring introduces high variance and hallucination noise that hinder optimization.

\noindent (2) \textbf{\textit{Superiority of Atomic Verification.}} Conversely, $R_{inq}$ alone outperforms $R_{judge}$. Combined with visual constraints ($R_{vis} + R_{inq}$), the model achieves SOTA results (High-Level: \textbf{91.6}, Low-Level: \textbf{81.7}). By decomposing evaluation into atomic, objective VQA tasks, \methodname{} ensures fidelity and mitigates the instability of holistic VLM scoring.

\subsection{Analysis}

\subsubsection{Reward Hacking Examination}

As shown in Figure~\ref{fig:train_curves}, the Train Reward, Pass Rate, and Consistency Reward improve consistently throughout the training phase.
To validate the efficacy of our RL stage and investigate potential reward hacking, where the model might maximize rewards by exploiting the execution rate at the expense of visual fidelity, we conduct an in-depth analysis of the training dynamics. 

\begin{figure}[t]
    \centering
    \includegraphics[width=1.0\columnwidth]{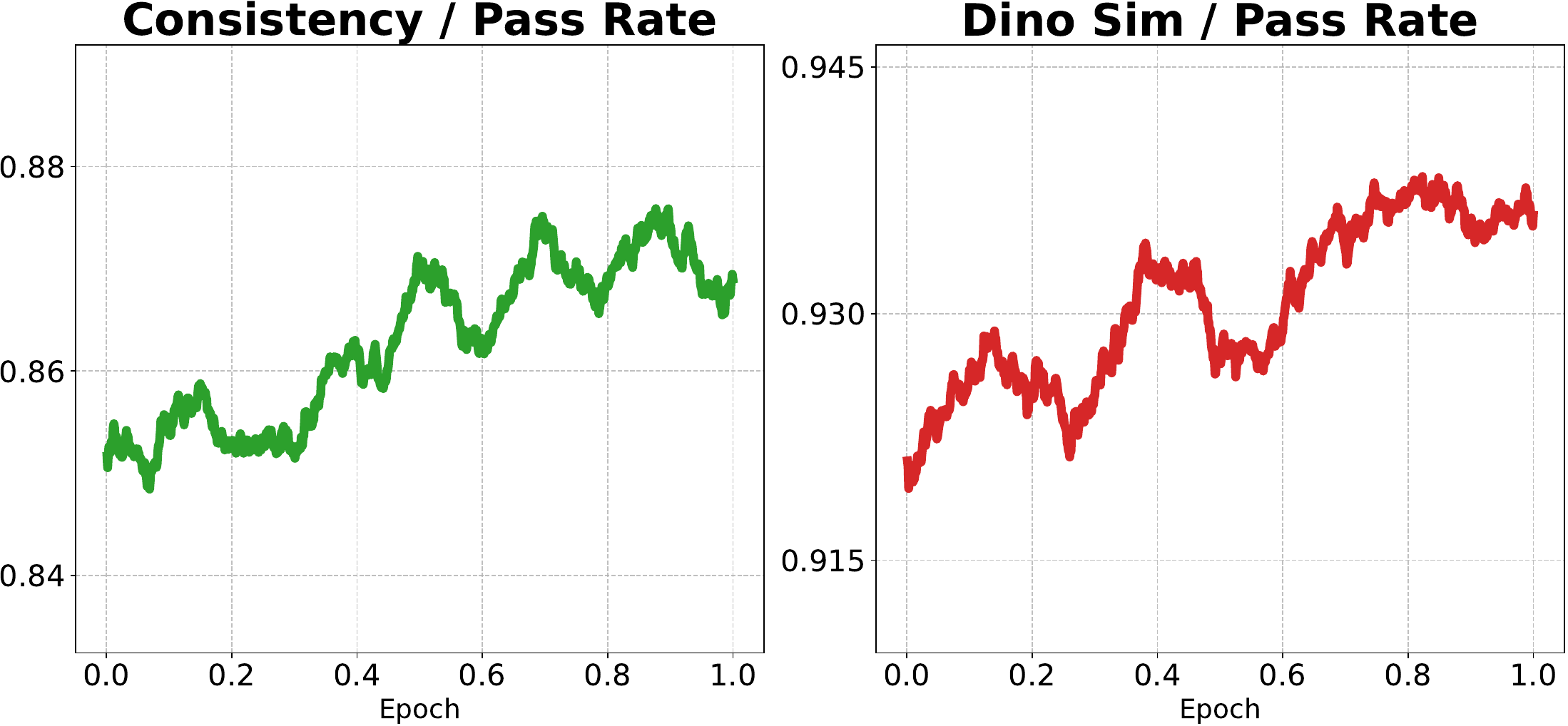}
    \caption{\textbf{Reward Hacking Analysis.} The continuous rise in rewards normalized by pass rate confirms that the model improves visual fidelity \textit{per executable sample}, distinct from merely exploiting execution rates.}
    \label{fig:ratio_curves}
\end{figure}

Specifically, we track the average reward per executable sample in Figure~\ref{fig:ratio_curves} to verify that the increase is not merely an artifact of generating more compilable code.
As illustrated by the \textbf{Consistency / Pass Rate} and \textbf{Visual Reward / Pass Rate} curves, even when conditioning on executed samples, the metrics exhibit a steady ascent over epochs.  
This implies a dual improvement: the model generates more executable code (\textbf{Quantity} $\uparrow$) while simultaneously achieving \textit{higher} visual fidelity within those valid samples (\textbf{Quality} $\uparrow$). 
This positive trend strongly refutes the reward hacking hypothesis, confirming that our Inquiry-Driven strategy successfully guides the model to optimize fine-grained visual grounding while ensuring syntactic correctness during the optimization process.

\begin{figure*}[ht]
    \centering
    \includegraphics[width=1.0\textwidth]{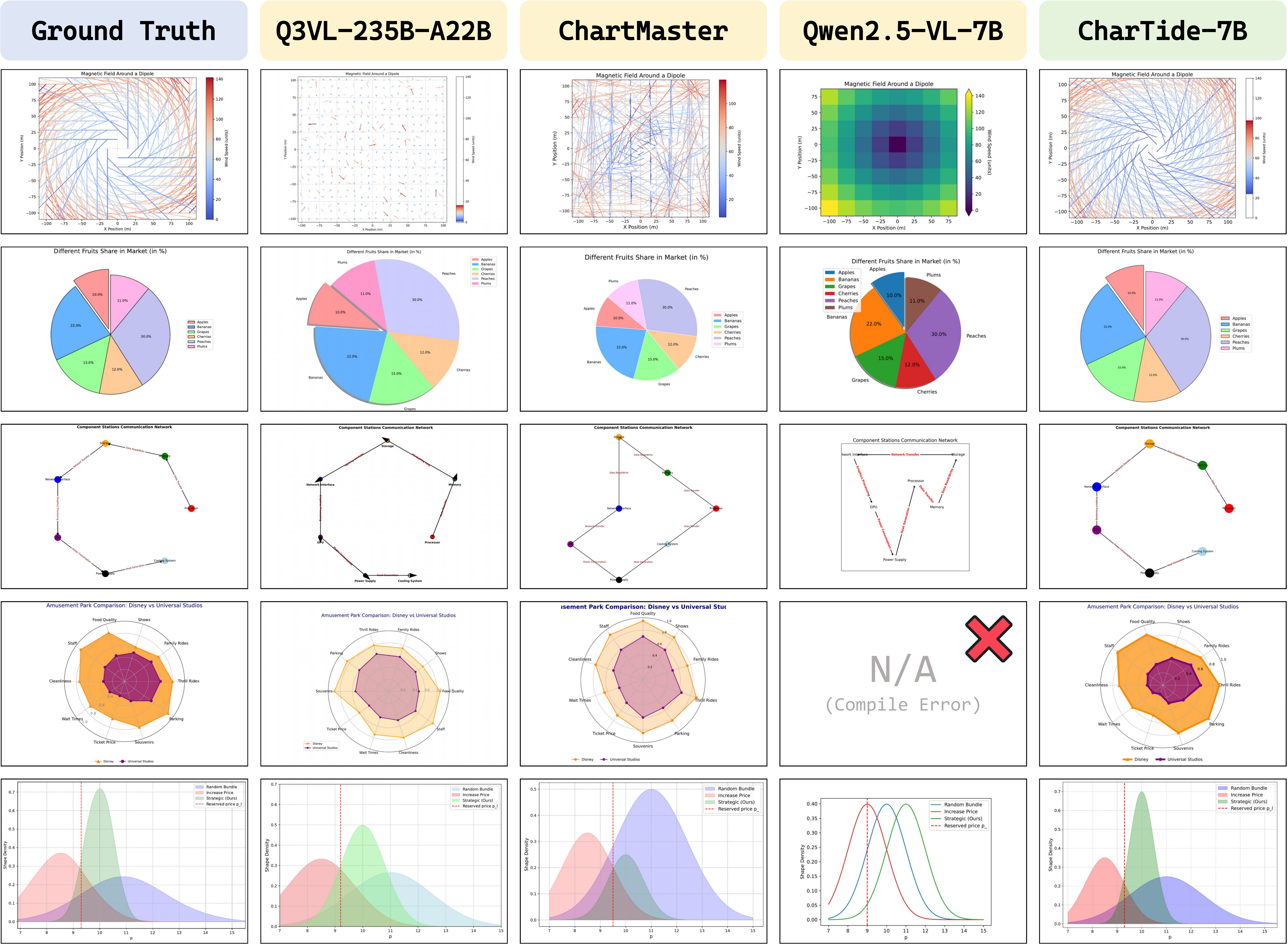}
    \small
    \caption{\textbf{Qualitative comparison on challenging chart types from ChartMimic Benchmark.} We compare \methodname{} against various baselines on topologically complex and information-dense samples. While baselines often suffer from structural collapse or omit fine-grained visual details, our model achieves high-fidelity reproduction aligned with the ground truth. We provide additional visualization examples in Appendix ~\ref{sec:appendix_more_visualizations}.}
    \label{fig:visualization}
\end{figure*}

\subsubsection{Data Contamination Analysis}

Although our Chart $\to$ Code training data is synthetically regenerated via Qwen3-VL-235B-A22B, theoretically preventing direct data leakage, we conduct a contamination analysis to rule out potential indirect leakage from the source ChartCap dataset. 
To ensure a robust assessment and mitigate the potential bias of a single feature extractor, we compute the cosine similarity based on the averaged feature embeddings from both WebSSL-1B and DINOv2-Giant encoders.
We perform this retrieval between every sample in the ChartMimic test set ($N=600$) and the entire 500k ChartCap dataset.
Detailed visualizations of the retrieval results are provided in Appendix~\ref{sec:appendix_leakage}. 

Our analysis reveals that even for the pairs with the highest averaged similarity scores, the retrieved training samples merely share similar chart types or color schemes but contain distinct data values and semantic contexts. 
This confirms that \methodname{} achieves performance gains through robust style learning and our tri-perspective training strategy, rather than by memorizing benchmarks.

\subsubsection{Qualitative Visualization}
Figure~\ref{fig:visualization} presents a qualitative comparison of complex charts, highlighting two key advantages:

\noindent \textbf{Structural Integrity.} 
For topologically complex charts such as quiver plots and graph structures, baseline models frequently exhibit structural hallucinations or incomplete generation. 
In contrast, \methodname{} faithfully reproduces vector field directions and intricate node connections, preserving the underlying topological logic.

\noindent \textbf{Detail Perception.} 
In dense scenarios like radar and density plots, baselines often generate rough outlines but struggle with fine-grained attributes like transparency and overlaps.
Conversely, \methodname{} maintains high fidelity in color mapping, data trends, and text annotations, corroborating the quantitative gains reported in Table~\ref{tab:main_results}. 

\vspace{-5pt}
\section{Conclusion}
\label{sec:conclusion}
\vspace{-5pt}

We propose \methodname{}, a unified data-centric framework for high-precision chart-to-code generation. 
By introducing the Tri-Perspective Tuning strategy, we effectively overcome the homogeneity bottleneck inherent in existing SFT data by explicitly decoupling visual perception from logical reasoning.
Furthermore, our Inquiry-Driven RL mechanism transforms the alignment paradigm, shifting from subjective VLM scoring to objective, atomic QA verification. This ensures rigorous consistency between visual semantics and the generated code.
Extensive experiments demonstrate that \methodname{} not only significantly outperforms SOTA open-source baselines but also achieves performance parity with top-tier proprietary models across authoritative benchmarks. 
We envision this work as a robust foundation for future advancements in more reliable multimodal code agents.

\section*{Limitations}
\label{sec:limitations}
Despite the comprehensive performance achieved by \methodname{} in resolving core challenges of chart-to-code generation, several avenues remain for future exploration. 
To ensure fair comparisons with mainstream open-source baselines, our experiments were primarily conducted within the 7B/8B parameter regime. While \methodname{} establishes a strong SOTA at this scale, the performance upper bounds accessible via larger backbones remain to be explored. 
Furthermore, our current evaluation concentrates on the \textit{Direct Mimic} task. 
Although the model demonstrates robust logic synthesis, we have not extensively evaluated scenarios involving iterative chart modification, style transfer, or refactoring based on customized user data. 
Future work will aim to extend \methodname{} to broader applications, such as interactive editing and cross-library type migration, bridging the gap between static reproduction and dynamic creation.


\section*{Ethical Considerations}
This work uses only publicly available datasets and open-source foundation models, following their specific licenses. Since the data is mostly scientific and synthetic charts, it contains no private personal details. Also, because the main goal is generating structured plotting code rather than general text, the risk of producing toxic or biased content is minimal. We believe this research helps make data visualization tools accessible to a wider audience, with no expected negative impact on society.


\bibliography{custom}

\begin{thebibliography}{41}
\providecommand{\natexlab}[1]{#1}

\bibitem[{Bai et~al.(2025{\natexlab{a}})Bai, Cai, Chen, Chen, Chen, Cheng, Deng, Ding, Gao, Ge, Ge, Guo, Huang, Huang, Huang, Hui, Jiang, Li, Li, Li, Li, Lin, Lin, Liu, Liu, Liu, Liu, Liu, Liu, Lu, Luo, Lv, Men, Meng, Ren, Ren, Song, Sun, Tang, Tu, Wan, Wang, Wang, Wang, Wang, Xie, Xu, Xu, Xu, Yang, Yang, Yang, Yang, Yu, Zhang, Zhang, Zhang, Zheng, Zhong, Zhou, Zhou, Zhou, Zhu, and Zhu}]{bai2025qwen3vltechnicalreport}
Shuai Bai, Yuxuan Cai, Ruizhe Chen, Keqin Chen, Xionghui Chen, Zesen Cheng, Lianghao Deng, Wei Ding, Chang Gao, Chunjiang Ge, Wenbin Ge, Zhifang Guo, Qidong Huang, Jie Huang, Fei Huang, Binyuan Hui, Shutong Jiang, Zhaohai Li, Mingsheng Li, and 45 others. 2025{\natexlab{a}}.
\newblock Qwen3-vl technical report.
\newblock \emph{arXiv preprint arXiv:2511.21631}.

\bibitem[{Bai et~al.(2025{\natexlab{b}})Bai, Chen, Liu, Wang, Ge, Song, Dang, Wang, Wang, Tang, Zhong, Zhu, Yang, Li, Wan, Wang, Ding, Fu, Xu, Ye, Zhang, Xie, Cheng, Zhang, Yang, Xu, and Lin}]{bai2025qwen25vltechnicalreport}
Shuai Bai, Keqin Chen, Xuejing Liu, Jialin Wang, Wenbin Ge, Sibo Song, Kai Dang, Peng Wang, Shijie Wang, Jun Tang, Humen Zhong, Yuanzhi Zhu, Mingkun Yang, Zhaohai Li, Jianqiang Wan, Pengfei Wang, Wei Ding, Zheren Fu, Yiheng Xu, and 8 others. 2025{\natexlab{b}}.
\newblock Qwen2.5-vl technical report.
\newblock \emph{arXiv preprint arXiv:2502.13923}.

\bibitem[{Chen et~al.(2025{\natexlab{a}})Chen, Zhao, Zeng, Huang, Zheng, Zhong, and Ma}]{chen2025breakingsftplateaumultimodal}
Lei Chen, Xuanle Zhao, Zhixiong Zeng, Jing Huang, Liming Zheng, Yufeng Zhong, and Lin Ma. 2025{\natexlab{a}}.
\newblock Breaking the sft plateau: Multimodal structured reinforcement learning for chart-to-code generation.
\newblock \emph{arXiv preprint arXiv:2508.13587}.

\bibitem[{Chen et~al.(2025{\natexlab{b}})Chen, Zhao, Zeng, Huang, Zhong, and Ma}]{chen2025chartr1}
Lei Chen, Xuanle Zhao, Zhixiong Zeng, Jing Huang, Yufeng Zhong, and Lin Ma. 2025{\natexlab{b}}.
\newblock \href {https://arxiv.org/abs/2507.15509} {Chart-r1: Chain-of-thought supervision and reinforcement for advanced chart reasoner}.
\newblock \emph{Preprint}, arXiv:2507.15509.

\bibitem[{Chen et~al.(2025{\natexlab{c}})Chen, Shen, Huang, Zhou, Lin, Cai, Yu, Bu, Shi, and Qiao}]{chen2025learningimagesvisualreinforcement}
Yang Chen, Yufan Shen, Wenxuan Huang, Sheng Zhou, Qunshu Lin, Xinyu Cai, Zhi Yu, Jiajun Bu, Botian Shi, and Yu~Qiao. 2025{\natexlab{c}}.
\newblock \href {https://arxiv.org/abs/2507.20766} {Learning only with images: Visual reinforcement learning with reasoning, rendering, and visual feedback}.
\newblock \emph{Preprint}, arXiv:2507.20766.

\bibitem[{DeepSeek-AI(2025)}]{deepseekai2025deepseekr1incentivizingreasoningcapability}
DeepSeek-AI. 2025.
\newblock \href {https://arxiv.org/abs/2501.12948} {Deepseek-r1: Incentivizing reasoning capability in llms via reinforcement learning}.
\newblock \emph{Preprint}, arXiv:2501.12948.

\bibitem[{Duan et~al.(2024)Duan, Yang, Qiao, Fang, Chen, Liu, Dong, Zang, Zhang, Wang et~al.}]{duan2024vlmevalkit}
Haodong Duan, Junming Yang, Yuxuan Qiao, Xinyu Fang, Lin Chen, Yuan Liu, Xiaoyi Dong, Yuhang Zang, Pan Zhang, Jiaqi Wang, and 1 others. 2024.
\newblock Vlmevalkit: An open-source toolkit for evaluating large multi-modality models.
\newblock In \emph{Proceedings of the 32nd ACM international conference on multimedia}, pages 11198--11201.

\bibitem[{Fan et~al.(2025)Fan, Tong, Zhu, Sinha, Liu, Chen, Rabbat, Ballas, LeCun, Bar et~al.}]{fan2025scalinglanguagefreevisualrepresentation}
David Fan, Shengbang Tong, Jiachen Zhu, Koustuv Sinha, Zhuang Liu, Xinlei Chen, Michael Rabbat, Nicolas Ballas, Yann LeCun, Amir Bar, and 1 others. 2025.
\newblock Scaling language-free visual representation learning.
\newblock \emph{arXiv preprint arXiv:2504.01017}.

\bibitem[{Gui et~al.(2025)Gui, Li, Wan, Shi, Zhang, Chen, Su, Chen, Wu, Zhou et~al.}]{gui2025webcode2m}
Yi~Gui, Zhen Li, Yao Wan, Yemin Shi, Hongyu Zhang, Bohua Chen, Yi~Su, Dongping Chen, Siyuan Wu, Xing Zhou, and 1 others. 2025.
\newblock Webcode2m: A real-world dataset for code generation from webpage designs.
\newblock In \emph{Proceedings of the ACM on Web Conference 2025}, pages 1834--1845.

\bibitem[{Han et~al.(2023)Han, Zhang, Chen, Yang, Wang, Yu, Fu, and Zhang}]{han2023chartllamamultimodalllmchart}
Yucheng Han, Chi Zhang, Xin Chen, Xu~Yang, Zhibin Wang, Gang Yu, Bin Fu, and Hanwang Zhang. 2023.
\newblock Chartllama: A multimodal llm for chart understanding and generation.
\newblock \emph{arXiv preprint arXiv:2311.16483}.

\bibitem[{He et~al.(2024)He, Xi, Zhao, Fan, Ding, Shan, Gui, Zhang, and Huang}]{he2025distillvisualchartreasoning}
Wei He, Zhiheng Xi, Wanxu Zhao, Xiaoran Fan, Yiwen Ding, Zifei Shan, Tao Gui, Qi~Zhang, and Xuanjing Huang. 2024.
\newblock Distill visual chart reasoning ability from llms to mllms.
\newblock \emph{arXiv preprint arXiv:2410.18798}.

\bibitem[{Huang et~al.(2025)Huang, Jia, Zhai, Cao, Ye, Zhao, Xu, Hu, and Lin}]{huang2025visionr1incentivizingreasoningcapability}
Wenxuan Huang, Bohan Jia, Zijie Zhai, Shaosheng Cao, Zheyu Ye, Fei Zhao, Zhe Xu, Yao Hu, and Shaohui Lin. 2025.
\newblock Vision-r1: Incentivizing reasoning capability in multimodal large language models.
\newblock \emph{arXiv preprint arXiv:2503.06749}.

\bibitem[{Jiang et~al.(2025{\natexlab{a}})Jiang, Huang, Wu, Li, Zhang, and Wei}]{jiang2025viscodex}
Lingjie Jiang, Shaohan Huang, Xun Wu, Yixia Li, Dongdong Zhang, and Furu Wei. 2025{\natexlab{a}}.
\newblock Viscodex: Unified multimodal code generation via merging vision and coding models.
\newblock \emph{arXiv preprint arXiv:2508.09945}.

\bibitem[{Jiang et~al.(2025{\natexlab{b}})Jiang, Zheng, Wan, Han, Wang, Lyu, and Yue}]{jiang2025screencoderadvancingvisualtocodegeneration}
Yilei Jiang, Yaozhi Zheng, Yuxuan Wan, Jiaming Han, Qunzhong Wang, Michael~R Lyu, and Xiangyu Yue. 2025{\natexlab{b}}.
\newblock Screencoder: Advancing visual-to-code generation for front-end automation via modular multimodal agents.
\newblock \emph{arXiv preprint arXiv:2507.22827}.

\bibitem[{Kondic et~al.(2025)Kondic, Li, Joshi, He, Abedin, Sun, Wiesel, Schwartz, Nassar, Wu et~al.}]{kondic2025chartgenscalingchartunderstanding}
Jovana Kondic, Pengyuan Li, Dhiraj Joshi, Zexue He, Shafiq Abedin, Jennifer Sun, Ben Wiesel, Eli Schwartz, Ahmed Nassar, Bo~Wu, and 1 others. 2025.
\newblock Chartgen: Scaling chart understanding via code-guided synthetic chart generation.
\newblock \emph{arXiv preprint arXiv:2507.19492}.

\bibitem[{Lambert et~al.(2024)Lambert, Morrison, Pyatkin, Huang, Ivison, Brahman, Miranda, Liu, Dziri, Lyu et~al.}]{lambert2025tulu3pushingfrontiers}
Nathan Lambert, Jacob Morrison, Valentina Pyatkin, Shengyi Huang, Hamish Ivison, Faeze Brahman, Lester James~V Miranda, Alisa Liu, Nouha Dziri, Shane Lyu, and 1 others. 2024.
\newblock Tulu 3: Pushing frontiers in open language model post-training.
\newblock \emph{arXiv preprint arXiv:2411.15124}.

\bibitem[{Lim et~al.(2025)Lim, Ahn, and Kim}]{lim2025chartcapmitigatinghallucinationdense}
Junyoung Lim, Jaewoo Ahn, and Gunhee Kim. 2025.
\newblock Chartcap: Mitigating hallucination of dense chart captioning.
\newblock In \emph{Proceedings of the IEEE/CVF International Conference on Computer Vision}, pages 13171--13182.

\bibitem[{Masry et~al.(2022)Masry, Do, Tan, Joty, and Hoque}]{masry2022chartqa}
Ahmed Masry, Xuan~Long Do, Jia~Qing Tan, Shafiq Joty, and Enamul Hoque. 2022.
\newblock Chartqa: A benchmark for question answering about charts with visual and logical reasoning.
\newblock In \emph{Findings of the association for computational linguistics: ACL 2022}, pages 2263--2279.

\bibitem[{Masry et~al.(2025)Masry, Islam, Ahmed, Bajaj, Kabir, Kartha, Laskar, Rahman, Rahman, Shahmohammadi et~al.}]{masry2025chartqapro}
Ahmed Masry, Mohammed~Saidul Islam, Mahir Ahmed, Aayush Bajaj, Firoz Kabir, Aaryaman Kartha, Md~Tahmid~Rahman Laskar, Mizanur Rahman, Shadikur Rahman, Mehrad Shahmohammadi, and 1 others. 2025.
\newblock Chartqapro: A more diverse and challenging benchmark for chart question answering.
\newblock In \emph{Findings of the Association for Computational Linguistics: ACL 2025}, pages 19123--19151.

\bibitem[{Meng et~al.(2025)Meng, Du, Liu, Zhou, Lu, Fu, Han, Shi, Wang, He et~al.}]{meng2025mmeurekaexploringfrontiersmultimodal}
Fanqing Meng, Lingxiao Du, Zongkai Liu, Zhixiang Zhou, Quanfeng Lu, Daocheng Fu, Tiancheng Han, Botian Shi, Wenhai Wang, Junjun He, and 1 others. 2025.
\newblock Mm-eureka: Exploring the frontiers of multimodal reasoning with rule-based reinforcement learning.
\newblock \emph{arXiv preprint arXiv:2503.07365}.

\bibitem[{Ni et~al.(2025)Ni, Yang, Li, Lin, Lin, Zuo, and Wang}]{ni2025point}
Minheng Ni, Zhengyuan Yang, Linjie Li, Chung-Ching Lin, Kevin Lin, Wangmeng Zuo, and Lijuan Wang. 2025.
\newblock Point-rft: Improving multimodal reasoning with visually grounded reinforcement finetuning.
\newblock \emph{arXiv preprint arXiv:2505.19702}.

\bibitem[{Niu et~al.(2025)Niu, Cui, Wang, Xu, Yao, Zhu, Wu, Wang, and Che}]{niu2025chart2code53}
Tianhao Niu, Yiming Cui, Baoxin Wang, Xiao Xu, Xin Yao, Qingfu Zhu, Dayong Wu, Shijin Wang, and Wanxiang Che. 2025.
\newblock Chart2code53: A large-scale diverse and complex dataset for enhancing chart-to-code generation.
\newblock In \emph{Proceedings of the 2025 Conference on Empirical Methods in Natural Language Processing}, pages 15839--15855.

\bibitem[{Oquab et~al.(2023)Oquab, Darcet, Moutakanni, Vo, Szafraniec, Khalidov, Fernandez, Haziza, Massa, El-Nouby et~al.}]{oquab2023dinov2}
Maxime Oquab, Timoth{\'e}e Darcet, Th{\'e}o Moutakanni, Huy Vo, Marc Szafraniec, Vasil Khalidov, Pierre Fernandez, Daniel Haziza, Francisco Massa, Alaaeldin El-Nouby, and 1 others. 2023.
\newblock Dinov2: Learning robust visual features without supervision.
\newblock \emph{arXiv preprint arXiv:2304.07193}.

\bibitem[{Shao et~al.(2024)Shao, Wang, Zhu, Xu, Song, Bi, Zhang, Zhang, Li, Wu et~al.}]{shao2024deepseekmathpushinglimitsmathematical}
Zhihong Shao, Peiyi Wang, Qihao Zhu, Runxin Xu, Junxiao Song, Xiao Bi, Haowei Zhang, Mingchuan Zhang, YK~Li, Yang Wu, and 1 others. 2024.
\newblock Deepseekmath: Pushing the limits of mathematical reasoning in open language models.
\newblock \emph{arXiv preprint arXiv:2402.03300}.

\bibitem[{Si et~al.(2025)Si, Zhang, Li, Yang, Liu, and Yang}]{si2025design2codebenchmarkingmultimodalcode}
Chenglei Si, Yanzhe Zhang, Ryan Li, Zhengyuan Yang, Ruibo Liu, and Diyi Yang. 2025.
\newblock Design2code: Benchmarking multimodal code generation for automated front-end engineering.
\newblock In \emph{Proceedings of the 2025 Conference of the Nations of the Americas Chapter of the Association for Computational Linguistics: Human Language Technologies (Volume 1: Long Papers)}, pages 3956--3974.

\bibitem[{Tan et~al.(2025)Tan, Cao, Xue, Zhan, Ding, and He}]{tan2025chartmasteradvancingcharttocodegeneration}
Wentao Tan, Qiong Cao, Chao Xue, Yibing Zhan, Changxing Ding, and Xiaodong He. 2025.
\newblock Chartmaster: Advancing chart-to-code generation with real-world charts and chart similarity reinforcement learning.
\newblock \emph{arXiv preprint arXiv:2508.17608}.

\bibitem[{Wan et~al.(2025)Wan, Wang, Dong, Wang, Li, Huo, and Lyu}]{Wan_2025}
Yuxuan Wan, Chaozheng Wang, Yi~Dong, Wenxuan Wang, Shuqing Li, Yintong Huo, and Michael Lyu. 2025.
\newblock Divide-and-conquer: Generating ui code from screenshots.
\newblock \emph{Proceedings of the ACM on Software Engineering}, 2(FSE):2099--2122.

\bibitem[{Wang et~al.(2025)Wang, Gao, Gu, Pu, Cui, Wei, Liu, Jing, Ye, Shao et~al.}]{wang2025internvl35advancingopensourcemultimodal}
Weiyun Wang, Zhangwei Gao, Lixin Gu, Hengjun Pu, Long Cui, Xingguang Wei, Zhaoyang Liu, Linglin Jing, Shenglong Ye, Jie Shao, and 1 others. 2025.
\newblock Internvl3.5: Advancing open-source multimodal models in versatility, reasoning, and efficiency.
\newblock \emph{arXiv preprint arXiv:2508.18265}.

\bibitem[{Wu et~al.(2025)Wu, Liang, Ge, Guo, Lu, Wang, Shan, and Luo}]{wu2024plot2codecomprehensivebenchmarkevaluating}
Chengyue Wu, Zhixuan Liang, Yixiao Ge, Qiushan Guo, Zeyu Lu, Jiahao Wang, Ying Shan, and Ping Luo. 2025.
\newblock Plot2code: A comprehensive benchmark for evaluating multi-modal large language models in code generation from scientific plots.
\newblock In \emph{Findings of the Association for Computational Linguistics: NAACL 2025}, pages 3006--3028.

\bibitem[{Xia et~al.(2025)Xia, Ye, Yan, Liu, Zhou, Chen, Shi, Yan, and Zhang}]{xia2025chartxchartvlmversatile}
Renqiu Xia, Hancheng Ye, Xiangchao Yan, Qi~Liu, Hongbin Zhou, Zijun Chen, Botian Shi, Junchi Yan, and Bo~Zhang. 2025.
\newblock Chartx \& chartvlm: A versatile benchmark and foundation model for complicated chart reasoning.
\newblock \emph{IEEE Transactions on Image Processing}.

\bibitem[{Xing et~al.(2025)Xing, Dong, Zang, Cao, Liang, Huang, Wang, Wu, and Lin}]{xing2025caprlstimulatingdenseimage}
Long Xing, Xiaoyi Dong, Yuhang Zang, Yuhang Cao, Jianze Liang, Qidong Huang, Jiaqi Wang, Feng Wu, and Dahua Lin. 2025.
\newblock Caprl: Stimulating dense image caption capabilities via reinforcement learning.
\newblock \emph{arXiv preprint arXiv:2509.22647}.

\bibitem[{Yang et~al.(2025{\natexlab{a}})Yang, Li, Yang, Zhang, Hui, Zheng, Yu, Gao, Huang, Lv, Zheng, Liu, Zhou, Huang, Hu, Ge, Wei, Lin, Tang, Yang, Tu, Zhang, Yang, Yang, Zhou, Zhou, Lin, Dang, Bao, Yang, Yu, Deng, Li, Xue, Li, Zhang, Wang, Zhu, Men, Gao, Liu, Luo, Li, Tang, Yin, Ren, Wang, Zhang, Ren, Fan, Su, Zhang, Zhang, Wan, Liu, Wang, Cui, Zhang, Zhou, and Qiu}]{yang2025qwen3technicalreport}
An~Yang, Anfeng Li, Baosong Yang, Beichen Zhang, Binyuan Hui, Bo~Zheng, Bowen Yu, Chang Gao, Chengen Huang, Chenxu Lv, Chujie Zheng, Dayiheng Liu, Fan Zhou, Fei Huang, Feng Hu, Hao Ge, Haoran Wei, Huan Lin, Jialong Tang, and 41 others. 2025{\natexlab{a}}.
\newblock Qwen3 technical report.
\newblock \emph{arXiv preprint arXiv:2505.09388}.

\bibitem[{Yang et~al.(2024)Yang, Shi, Liu, Shui, Wang, Jing, Xu, Zhu, Li, Zhang et~al.}]{yang2025chartmimicevaluatinglmmscrossmodal}
Cheng Yang, Chufan Shi, Yaxin Liu, Bo~Shui, Junjie Wang, Mohan Jing, Linran Xu, Xinyu Zhu, Siheng Li, Yuxiang Zhang, and 1 others. 2024.
\newblock Chartmimic: Evaluating lmm's cross-modal reasoning capability via chart-to-code generation.
\newblock \emph{arXiv preprint arXiv:2406.09961}.

\bibitem[{Yang et~al.(2025{\natexlab{b}})Yang, He, Pan, Jiang, Deng, Yang, Lu, Yin, Rao, Zhu, Zhang, and Chen}]{yang2025r1onevision}
Yi~Yang, Xiaoxuan He, Hongkun Pan, Xiyan Jiang, Yan Deng, Xingtao Yang, Haoyu Lu, Dacheng Yin, Fengyun Rao, Minfeng Zhu, Bo~Zhang, and Wei Chen. 2025{\natexlab{b}}.
\newblock R1-onevision: Advancing generalized multimodal reasoning through cross-modal formalization.
\newblock \emph{arXiv preprint arXiv:2503.10615}.

\bibitem[{Yun et~al.(2024)Yun, Thushara, Bhat, Wang, Deng, Wang, Tao, Li, Li, Nakov et~al.}]{yun2024web2codelargescalewebpagetocodedataset}
Sukmin Yun, Rusiru Thushara, Mohammad Bhat, Yongxin Wang, Mingkai Deng, Jinhong Wang, Tianhua Tao, Junbo Li, Haonan Li, Preslav Nakov, and 1 others. 2024.
\newblock Web2code: A large-scale webpage-to-code dataset and evaluation framework for multimodal llms.
\newblock \emph{Advances in neural information processing systems}, 37:112134--112157.

\bibitem[{Zha et~al.(2025)Zha, Zhou, Wu, Wang, Feng, Xu, Hao, Liu, Xing, and Hu}]{zha2025vision}
Yuheng Zha, Kun Zhou, Yujia Wu, Yushu Wang, Jie Feng, Zhi Xu, Shibo Hao, Zhengzhong Liu, Eric~P Xing, and Zhiting Hu. 2025.
\newblock Vision-g1: Towards general vision language reasoning with multi-domain data curation.
\newblock \emph{arXiv preprint arXiv:2508.12680}.

\bibitem[{Zhai et~al.(2023)Zhai, Mustafa, Kolesnikov, and Beyer}]{zhai2023sigmoid}
Xiaohua Zhai, Basil Mustafa, Alexander Kolesnikov, and Lucas Beyer. 2023.
\newblock Sigmoid loss for language image pre-training.
\newblock In \emph{Proceedings of the IEEE/CVF international conference on computer vision}, pages 11975--11986.

\bibitem[{Zhang et~al.(2025{\natexlab{a}})Zhang, Zuo, He, Sun, Liu, Jiang, Fan, Tian, Jia, Li et~al.}]{zhang2025survey}
Kaiyan Zhang, Yuxin Zuo, Bingxiang He, Youbang Sun, Runze Liu, Che Jiang, Yuchen Fan, Kai Tian, Guoli Jia, Pengfei Li, and 1 others. 2025{\natexlab{a}}.
\newblock A survey of reinforcement learning for large reasoning models.
\newblock \emph{arXiv preprint arXiv:2509.08827}.

\bibitem[{Zhang et~al.(2025{\natexlab{b}})Zhang, Cao, and Liao}]{zhang2025boostingcharttocodegenerationmllm}
Zhihan Zhang, Yixin Cao, and Lizi Liao. 2025{\natexlab{b}}.
\newblock Boosting chart-to-code generation in mllm via dual preference-guided refinement.
\newblock In \emph{Proceedings of the 33rd ACM International Conference on Multimedia}, pages 11032--11041.

\bibitem[{Zhao et~al.(2025{\natexlab{a}})Zhao, Jiang, Zeng, Chen, Qiu, Huang, Zhong, Zheng, Cao, and Ma}]{zhao2025vincicoderunifyingmultimodalcode}
Xuanle Zhao, Deyang Jiang, Zhixiong Zeng, Lei Chen, Haibo Qiu, Jing Huang, Yufeng Zhong, Liming Zheng, Yilin Cao, and Lin Ma. 2025{\natexlab{a}}.
\newblock Vincicoder: Unifying multimodal code generation via coarse-to-fine visual reinforcement learning.
\newblock \emph{arXiv preprint arXiv:2511.00391}.

\bibitem[{Zhao et~al.(2025{\natexlab{b}})Zhao, Luo, Shi, Chen, Wang, Liu, and Sun}]{zhao2025chartcoderadvancingmultimodallarge}
Xuanle Zhao, Xianzhen Luo, Qi~Shi, Chi Chen, Shuo Wang, Zhiyuan Liu, and Maosong Sun. 2025{\natexlab{b}}.
\newblock Chartcoder: Advancing multimodal large language model for chart-to-code generation.
\newblock \emph{arXiv preprint arXiv:2501.06598}.

\end{thebibliography}

\appendix
\clearpage



\section{Evaluation Details}
\label{sec:appendix_eval}

To ensure a strictly fair comparison, all results reported in Table~\ref{tab:main_results} were re-evaluated using the official open-source weights under identical experimental settings. We did not rely on reported numbers from original papers, which may vary due to differences in evaluation environments or API versions. The specific protocols for each benchmark are as follows:

\vspace{0.5em}
\noindent \textbf{ChartMimic.} 
We utilize the test set from the \textbf{ChartMimic v2} release~\cite{yang2025chartmimicevaluatinglmmscrossmodal}, specifically the ``Direct Mimic'' task, which consists of 600 chart images. We conduct evaluations using the \texttt{vlmevalkit}~\cite{duan2024vlmevalkit} framework, adhering strictly to its default settings. The evaluation metrics consist of:
(1) \textbf{High-Level Score (0--100):} Assessed by \texttt{GPT-4o-2024-05-13} to measure overall visual and semantic consistency;
(2) \textbf{Low-Level Score:} Calculated as the average F1 score across four atomic dimensions: text, layout, chart type, and color. Results are reported as the mean of three independent measurements per test.

\vspace{0.5em}
\noindent \textbf{Plot2Code.} 
We employ the official evaluation pipeline~\cite{wu2024plot2codecomprehensivebenchmarkevaluating} on the standard test set comprising 133 scientific plots. Metrics include Code Execution Rate, Text Match, and Visual Similarity Rating.
\textit{Note on Re-evaluation:} 
(1) \textbf{Judge Model Update:} Since the original judge model, GPT-4V, has been deprecated, we replaced it with \textbf{GPT-4o} for the Visual Similarity Rating. Consequently, we re-ran evaluations for all baseline models using this upgraded judge to ensure a unified and fair comparison.
(2) \textbf{Metric Normalization Adjustment:} The original Plot2Code benchmark computes Text Match and Rating scores only on \textit{executable} samples, which artificially inflates the visual scores for models with low execution rates (survivorship bias). To address this, we modified the evaluation to calculate the average score over the \textbf{total number of test samples} (assigning zero to failed executions). This rigorous standard ensures a holistic assessment but may result in numerical discrepancies compared to prior literature.

\vspace{0.5em}
\noindent \textbf{ChartX.} 
We focus on the \textbf{Chart Generation/Reproduction} task within the ChartX benchmark~\cite{xia2025chartxchartvlmversatile}, which contains 1,152 chart images. Performance is measured using a GPT-4 based scoring metric (on a scale of 0--5), which evaluates the semantic alignment between the generated code logic and the visual attributes of the ground truth.

\subsection{OOD Generalization Beyond Direct Mimic}
\label{sec:appendix_ood}

\noindent \textbf{Customized Mimic.}
We additionally evaluate zero-shot transfer on the Customized Mimic split of ChartMimic using the same evaluation toolkit and default settings as the Direct Mimic protocol.
This setting is more challenging because the model must follow less literal reproduction instructions without any task-specific fine-tuning.

\noindent \textbf{Seaborn OOD.}
To test backend generalization, we build a small OOD set from 47 official Seaborn example scripts.
Because some low-level metrics in ChartMimic assume a matplotlib execution stack, we report execution rate, average score over all examples, and average score over executable examples.
We use GPT-4o-2024-11-20 as the judge model for the high-level evaluation.

\begin{table}[h]
    \centering
    \caption{\textbf{Detailed OOD evaluation results.}}
    \label{tab:appendix_ood}
    \resizebox{\columnwidth}{!}{
    \begin{tabular}{llccc}
        \toprule
        \textbf{Setting} & \textbf{Model} & \textbf{Exec.} & \textbf{Avg(all)} & \textbf{Avg(exec)} \\
        \midrule
        \multirow{2}{*}{Customized Mimic} & Qwen2.5-VL-7B & 79.0 & 56.42 & 64.61 \\
         & \methodname{}-7B & \textbf{92.7} & \textbf{62.91} & \textbf{79.09} \\
        \midrule
        \multirow{2}{*}{Seaborn OOD} & Qwen2.5-VL-7B & 61.70 & 41.04 & 66.52 \\
         & \methodname{}-7B & \textbf{82.98} & \textbf{69.40} & \textbf{83.64} \\
        \bottomrule
    \end{tabular}}
\end{table}

\subsection{Significance Test for Inquiry-Driven Reward}
\label{sec:appendix_significance}

We perform a paired t-test over the 600 ChartMimic test samples, comparing $R_{vis}+R_{judge}$ against $R_{vis}+R_{inq}$.

\begin{table}[h]
    \centering
    \caption{\textbf{Paired significance test for the proposed reward.}}
    \label{tab:appendix_significance}
    \resizebox{\columnwidth}{!}{
    \begin{tabular}{lccc}
        \toprule
        \textbf{Metric} & $\boldsymbol{\Delta}$ \textbf{Mean} & \textbf{$p$-value} & \textbf{Interpretation} \\
        \midrule
        Overall & +1.41 & 0.027 & Significant \\
        Chart Type & +2.53 & 0.019 & Significant \\
        Layout & +1.58 & 0.055 & Marginal \\
        Color & +1.77 & 0.086 & Marginal \\
        \bottomrule
    \end{tabular}}
\end{table}

In addition, 12.8\% of the test samples (77/600) improve by more than 10 points, indicating that the gain is not merely a uniform minor shift.

\section{SFT Data Source and Pipeline Details}
\label{sec:appendix_data}

The core foundation of our training data is derived from \textbf{ChartCap}~\cite{lim2025chartcapmitigatinghallucinationdense}, a large-scale dataset containing 500k high-quality charts paired with detailed captions and info. We process this data into three distinct streams to support our Tri-Perspective SFT strategy.

\subsection{Visual Perception Stream (Chart $\to$ Caption)}
For the visual perception stream, we directly leverage the image-caption pairs from ChartCap. To ensure compatibility with the context window of our base models and to remove potential noise from excessively long text, we apply a length-based filter, excluding samples where the caption length exceeds 4,096 tokens. This results in a high-quality dataset focused on dense visual description.

\subsection{Code Logic Stream (Caption $\to$ Code)}
To bolster the model's ability to synthesize code from pure textual logic, we construct a \textit{Caption-to-Code} dataset.
\paragraph{Synthesis.} We concatenate the descriptive caption and the structural `info` (data table summary) from ChartCap to form a rich textual context. We then employ \textbf{Qwen3-Coder-30B-A3B} to generate the corresponding Python plotting code. The prompt used for this generation is as follows:

\begin{tcolorbox}[breakable, enhanced, colback=white, colframe=black, arc=0pt, outer arc=0pt, boxrule=0.8pt, left=2mm, right=2mm, top=2mm, bottom=2mm]
    \textbf{System Prompt:} You are an expert Python data visualization engineer. \\
    \textbf{User Instruction:} Based on the following chart description and data summary, please write Python code using Matplotlib/Seaborn to reproduce the chart. \\
    \textbf{Input Data:} \\
    \texttt{[Caption]:} \{input\_caption\} \\
    \texttt{[Data Info]:} \{input\_info\} \\
    \textbf{Constraints:} Ensure the code is self-contained, executable, and strictly follows the data trends described.
\end{tcolorbox}

\paragraph{Filtering Pipeline.} To ensure data quality, we execute the generated code to render a new chart image ($I_{gen}$). We then employ a rigorous consistency check using \textbf{Qwen3-VL-235B-A22B}. Specifically, we feed the original ground-truth image ($I_{src}$), the generated image ($I_{gen}$), and the original caption into the VLM, instructing it to evaluate whether $I_{gen}$ semantically and visually aligns with $I_{src}$ given the text description. After filtering out execution failures and low-consistency samples, we retain approximately \textbf{400k} high-quality pairs.

\subsection{Modality Fusion Stream (Chart $\to$ Code)}
This stream targets the end-to-end capability of converting visual charts directly into code.
\paragraph{Data Aggregation.} We expand the data scale by combining the 500k images from ChartCap with an additional 500k charts collected from open-source datasets used in prior works~\cite{zhao2025chartcoderadvancingmultimodallarge, zhao2025vincicoderunifyingmultimodalcode}(with Apache-2.0 license). This yields a total pool of \textbf{1 million} chart images.

\paragraph{Code Generation.} We utilize the powerful \textbf{Qwen3-VL-235B-A22B} model to generate ground-truth quality code for these images. The prompt is designed to enforce strict visual alignment:

\begin{tcolorbox}[breakable, enhanced, colback=white, colframe=black, arc=0pt, outer arc=0pt, boxrule=0.8pt, left=2mm, right=2mm, top=2mm, bottom=2mm]
\noindent
    \textbf{System Prompt:} You are an expert in chart reverse engineering. \\
    \textbf{User Instruction:} Write Python code to reproduce this chart image using matplotlib. \\
    \textbf{Input Image:} [Image Placeholder] \\
    \textbf{Requirements:}
    \begin{itemize}[left=0.2em]
\setlength{\itemsep}{0.05em}    
\setlength{\parskip}{0.1em}
        \item[1.] The code must be executable.
        \item[2.] Use the exact data values visible in the chart.
        \item[3.] Strictly mimic the styling (colors, fonts, legend position, markers).
    \end{itemize}
  
\end{tcolorbox}

Post-generation, we execute the code to render the predicted charts ($I_{pred}$). To eliminate visual hallucinations and ensure high reproduction fidelity, we implement a strict visual verification mechanism. We utilize the \textbf{WebSSL-1B}\cite{fan2025scalinglanguagefreevisualrepresentation} encoder to extract visual feature embeddings from both the original ground-truth chart ($I_{src}$) and the rendered chart ($I_{pred}$). We then compute the cosine similarity between these embeddings and retain only those samples where the similarity score exceeds a threshold of \textbf{0.8}. This rigorous filtering process ensures that the final training data maintains high visual alignment standards.

\subsection{Information Asymmetry Analysis}
\label{sec:appendix_asymmetry}

To quantify the supervision imbalance in end-to-end Chart$\to$Code training, we randomly sample 5,000 Chart$\to$Code examples (approximately 2.33M tokens in total) and categorize tokens into: (1) boilerplate template, (2) data definition, (3) visual configuration, (4) plotting calls, and (5) other code.
We then further isolate the subset of visual attribute values that truly need to be inferred from the image, such as color, marker, fontsize, linewidth, linestyle, and alpha.

\begin{table}[h]
    \centering
    \caption{\textbf{Token composition of Chart$\to$Code supervision.}}
    \label{tab:token_asymmetry}
    \resizebox{\columnwidth}{!}{
    \begin{tabular}{lcl}
        \toprule
        \textbf{Category} & \textbf{Share} & \textbf{Description} \\
        \midrule
        Non-visual tokens & 70.8\% & imports, figure init, comments, data processing \\
        Visual-related tokens & 29.2\% & titles, legends, ticks, plotting APIs \\
        Visual attribute values & 2.9\% & color, marker, fontsize, linewidth, linestyle, alpha \\
        \bottomrule
    \end{tabular}}
\end{table}

The ratio between non-visual tokens and truly image-conditioned visual-attribute values is therefore approximately 25:1.
Moreover, for several visual attributes (e.g., marker, fontsize, linewidth, and linestyle), the Top-3 values cover 63\%--86\% of occurrences.
This confirms that standard cross-entropy training is naturally dominated by predictable scaffolding and common-value patterns, which encourages template memorization over fine-grained visual grounding.

\section{QA Data Generation Details}
\label{sec:appendix_qa_gen}

To construct a high-quality, verifiable reward model for the RL stage, we curated a specialized VQA dataset. The pipeline consists of three phases: representative sampling, question generation, and consistency filtering.

\subsection{Representative Sampling}
Given the redundancy in the large-scale pre-training data, we aim to select a diverse subset for reinforcement learning. We first utilize the \textbf{WebSSL-1B} encoder to extract high-dimensional visual feature vectors from the 500k charts in the ChartCap dataset. Subsequently, we apply K-Means clustering on these features to identify distinct visual clusters. From these clusters, we sample \textbf{30k} representative chart images, ensuring coverage of diverse chart types, layouts, and data distributions.

\subsection{Question Generation}
For each selected chart, we leverage \textbf{GPT-5} to generate a set of 10 evaluation questions ($N=10$). To ensure the ground truth answers are accurate, we provide GPT-5 with both the rich semantic captions and the underlying data tables (info) from ChartCap as context. The system prompt used to guide GPT-5 is designed to cover multiple dimensions of chart assessment:

\begin{tcolorbox}[breakable, enhanced, colback=white, colframe=black, arc=0pt, outer arc=0pt, boxrule=0.8pt, left=2mm, right=2mm, top=2mm, bottom=2mm]
    \small
    \textbf{System Prompt:} You are a Data Visualization Benchmark Creator. Your task is to generate \textbf{10 Evaluation Questions} to judge if a reconstructed chart matches the original.

    \vspace{0.3em}
    \textbf{DATA SOURCE RULES:}
    \begin{enumerate}[left=0.2em]
\setlength{\itemsep}{0.05em}    
\setlength{\parskip}{0.1em}
        \item \textbf{DATA \& TEXT:} TRUST THE CAPTION. If caption says ``16\%'', answer is 16.
        \item \textbf{STYLE \& LAYOUT:} LOOK AT THE IMAGE for visual details missing in text.
    \end{enumerate}

    \textbf{DIFFICULTY SETTINGS:}
    \begin{itemize}[left=0.2em]
\setlength{\itemsep}{0.05em}    
\setlength{\parskip}{0.1em}
        \item \textbf{Keep it Simple:} Focus on obvious data retrieval (e.g., ``What is the max value?'') rather than complex math.
        \item \textbf{Generous Tolerance:} For numerical questions, allow a wider margin of error (e.g., if value is 50, tolerance should be 5.0). The goal is to check if the trend is correct, not pixel-perfect precision.
    \end{itemize}

    \textbf{QUESTION DISTRIBUTION (Total 10):}
    \begin{itemize}[left=0.2em]
\setlength{\itemsep}{0.05em}    
\setlength{\parskip}{0.1em}
        \item \textbf{Chart Type (1 Q - Bool):} Is it the correct chart type? (e.g., ``Is this a stacked bar chart?'')
        \item \textbf{Layout (1 Q - Bool):} Check subplot arrangement or legend position.
        \item \textbf{Text Content (3 Qs - Bool):}
            \begin{itemize}[left=0.2em]
\setlength{\itemsep}{0.05em}    
\setlength{\parskip}{0.1em}
                \item 2 Positive checks (e.g., ``Is X-axis label `Year'?'').
                \item 1 Negative/Trick check (e.g., ``Does title contain `2050'?'' $\to$ No).
            \end{itemize}
        \item \textbf{Data Accuracy (3 Qs - Float):}
            \begin{itemize}[left=0.2em]
\setlength{\itemsep}{0.05em}    
\setlength{\parskip}{0.1em}
                \item Ask for 3 distinct data values mentioned in the caption.
                \item \textbf{CRITICAL:} Set a `tolerance` of at least 5-10\% of the value.
            \end{itemize}
        \item \textbf{Style (2 Qs - Bool):} Check Colors or Marker shapes (e.g., ``Are bars red?'', ``Is line dashed?'').
    \end{itemize}
    
    \textbf{User Input:} Ground Truth Caption: \{caption\} \\
    Generate the 10 QA pairs now.
\end{tcolorbox}

\subsection{Consistency Filtering}
To mitigate potential hallucinations in the generation process and ensure the questions are answerable by a visual inspector, we implement a validation step. We employ \textbf{Qwen3-VL-30B-A3B} to answer the generated questions based on the original chart images. We verify the model's answers against the GPT-5 generated ground truth, applying the specified numerical tolerance for data-related questions. We filter out any charts where the model fails to answer correctly, retaining approximately \textbf{20k} high-quality, verified QA samples for the final RL training set.

\section{More Qualitative Analysis}

\subsection{Qualitative Analysis of Visual Models}
\label{sec:appendix_webssl}

\begin{figure}[h]
    \centering
    \includegraphics[width=1.0\columnwidth]{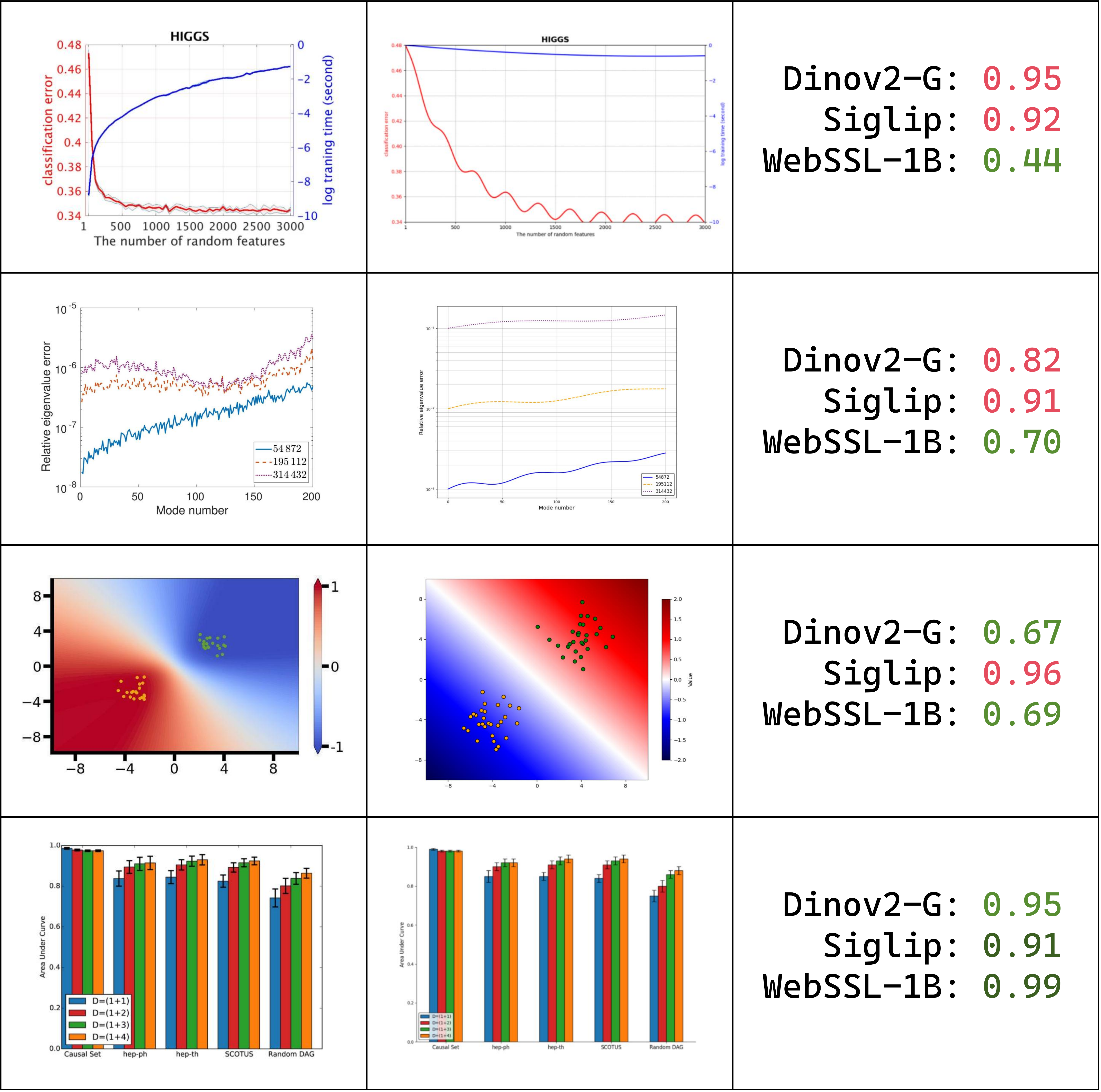} 
    \caption{\textbf{Qualitative comparison of visual similarity matching.} We display sample pairs from the SFT filtering stage. The scores assigned by WebSSL-1B (shown above each pair) correlate strongly with the visual and structural consistency observed by human judges, effectively identifying high-quality reproductions while rejecting structural hallucinations.}
    \label{fig:webssl_qualitative}
    \vspace{-10pt}
\end{figure}

As illustrated in Figure~\ref{fig:webssl_qualitative}, to intuitively validate the superiority of WebSSL-1B in assessing chart-to-code reconstruction quality, we qualitatively analyzed representative pairs from the Modality Fusion filtering pipeline.

We observed that standard visual encoders often fail to penalize subtle structural errors or semantic deviations if the overall color histogram remains similar. In contrast, WebSSL-1B demonstrates remarkable alignment with human perception by strictly penalizing structural collapses and fine-grained visual discrepancies (e.g., marker shapes, line styles), while implicitly capturing data trend deviations to effectively filter out ``plausible but wrong'' samples. These characteristics confirm that WebSSL-1B serves as a robust discriminative filter for our data construction and RL training.

\subsection{Data Contamination Analysis}
\label{sec:appendix_leakage}

To ensure that our model's high performance stems from learning generalized plotting rules rather than simply memorizing the training data, we conducted a thorough data contamination check.

We computed the visual similarity between every image in the ChartMimic test set ($N=600$) and the ChartCap 500k training dataset. To ensure a robust assessment, we employed two distinct encoders: \textbf{WebSSL-1B} and \textbf{DINOv2-Giant}. Specifically, we calculated the cosine similarity scores independently for each encoder and then averaged these scores to determine the final ranking. Based on these averaged scores, we retrieved the most similar training samples for each test case to identify potential leakage.

\begin{figure*}[ht]
    \centering
    \includegraphics[width=1.0\textwidth]{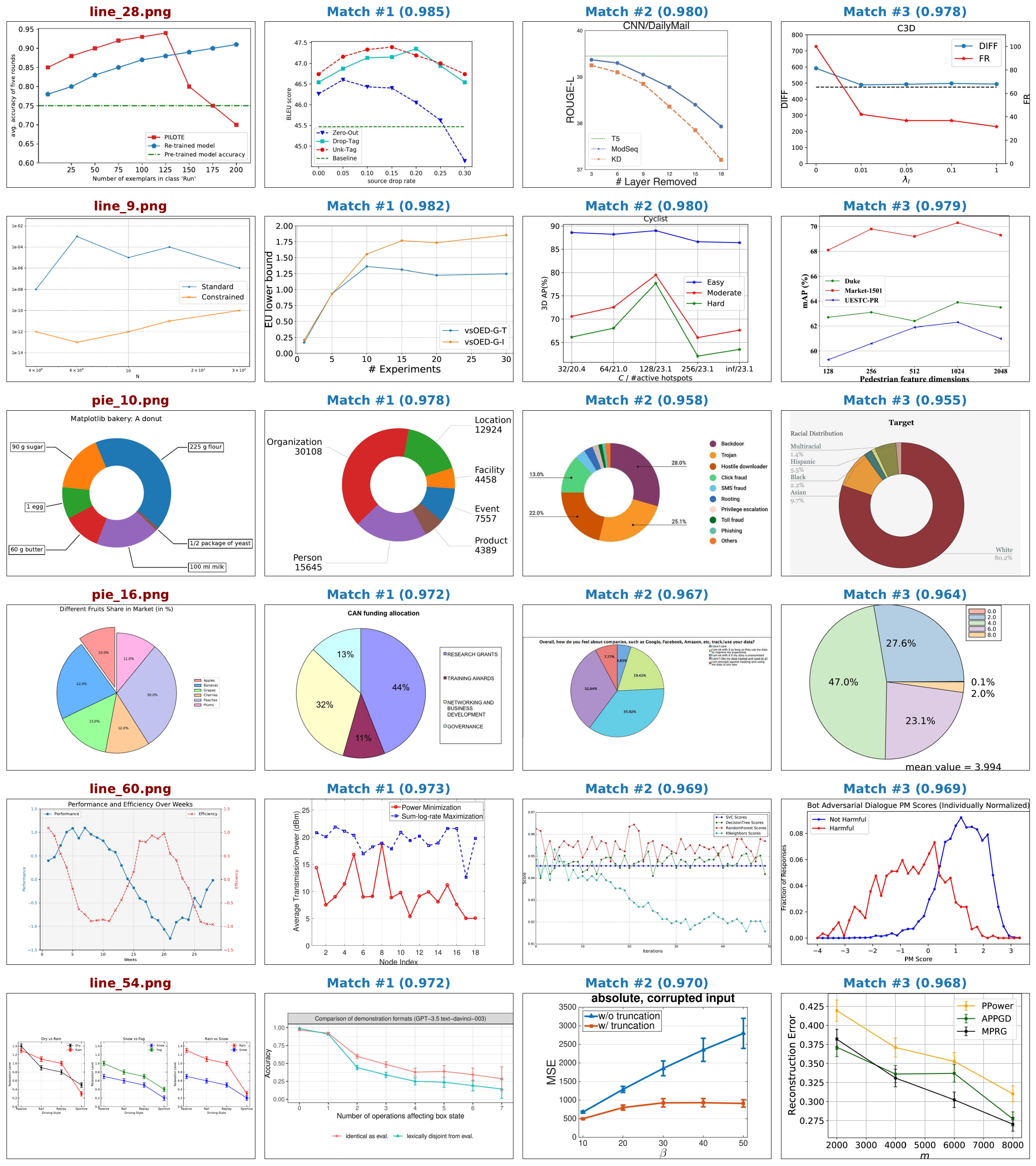} 
    \caption{\textbf{Worst-case leakage check.} We display the 6 ChartMimic test samples with the \textit{highest} similarity scores against the chartcap training set. The retrieved training samples (right) visually resemble the queries (left) in style and layout but differ in specific content and data values, confirming no direct leakage exists.}
    \label{fig:leakage_top}
\end{figure*}

Figure~\ref{fig:leakage_top} presents the ``worst-case'' scenario, showing the 6 test images that obtained the highest averaged similarity scores with the training set.
We observe that while these pairs share a high degree of stylistic similarity, such as using standard Matplotlib color palettes, default layouts, or identical chart types, they are not duplicates.
Notably, the actual content, including data values, axis ranges, and textual labels, remains different. This indicates that while the model has seen similar styles, it has not seen the specific data used in the test set.

\begin{figure*}[ht]
    \centering
    \includegraphics[width=1.0\textwidth]{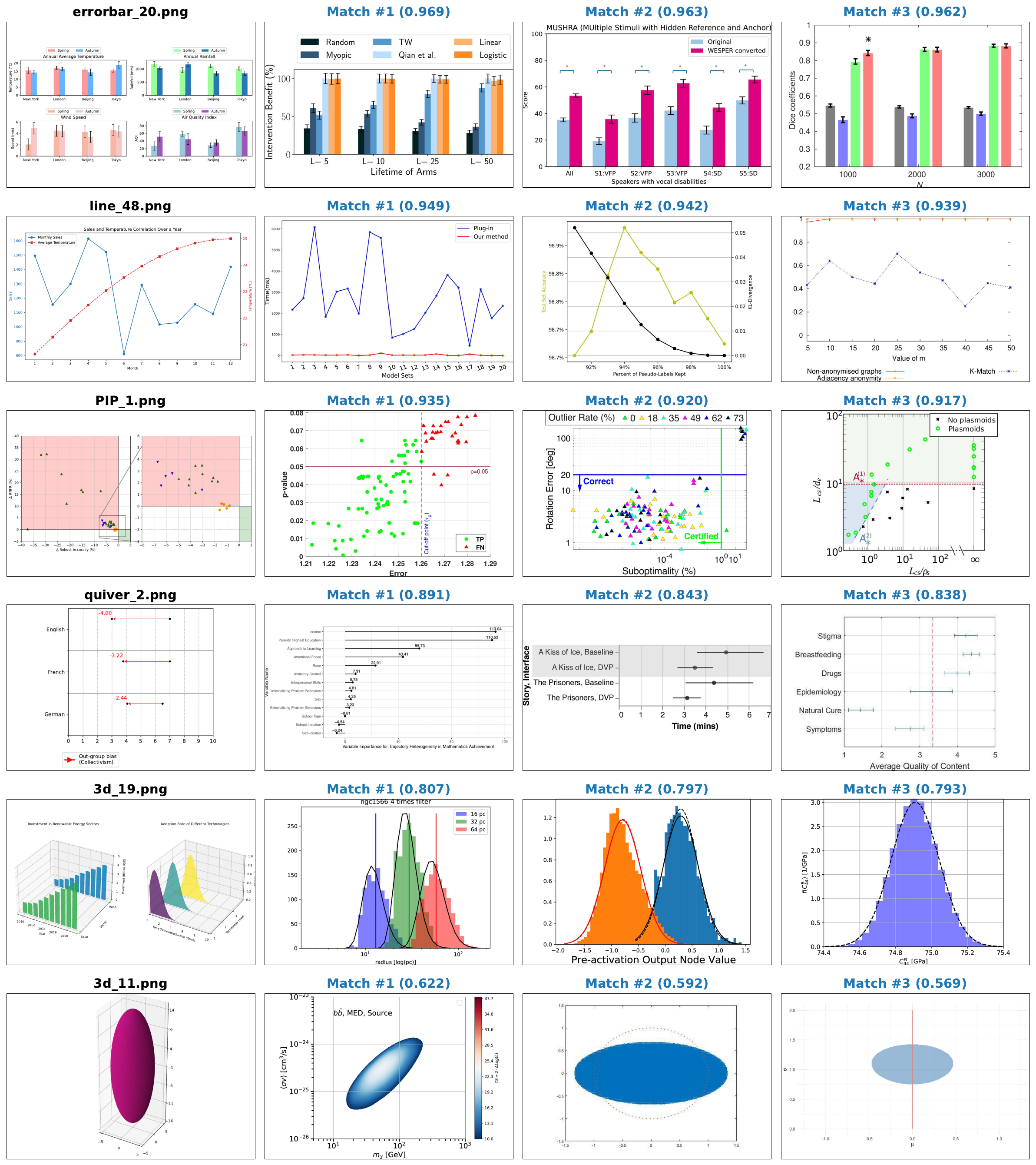}
    \caption{\textbf{Average-case leakage check.} We visualize 6 uniformly sampled ChartMimic images and their nearest neighbors in the chartcap training set, showing significant semantic and visual differences.}
    \label{fig:leakage_random}
\end{figure*}

For a broader perspective, Figure~\ref{fig:leakage_random} shows 6 randomly selected test images and their nearest training neighbors based on the averaged metric.
In this average case, the retrieved training images are visually and semantically distinct from the test queries, confirming that the vast majority of the test set has no close counterparts in the training data.
These results confirm that \methodname{} achieves its performance through style generalization and robust instruction following, rather than data leakage.

\clearpage
\onecolumn

\subsection{More Visualizations}
\label{sec:appendix_more_visualizations}

\vspace{20pt}

\begin{figure}[h]
    \centering
    \includegraphics[width=1.0\textwidth]{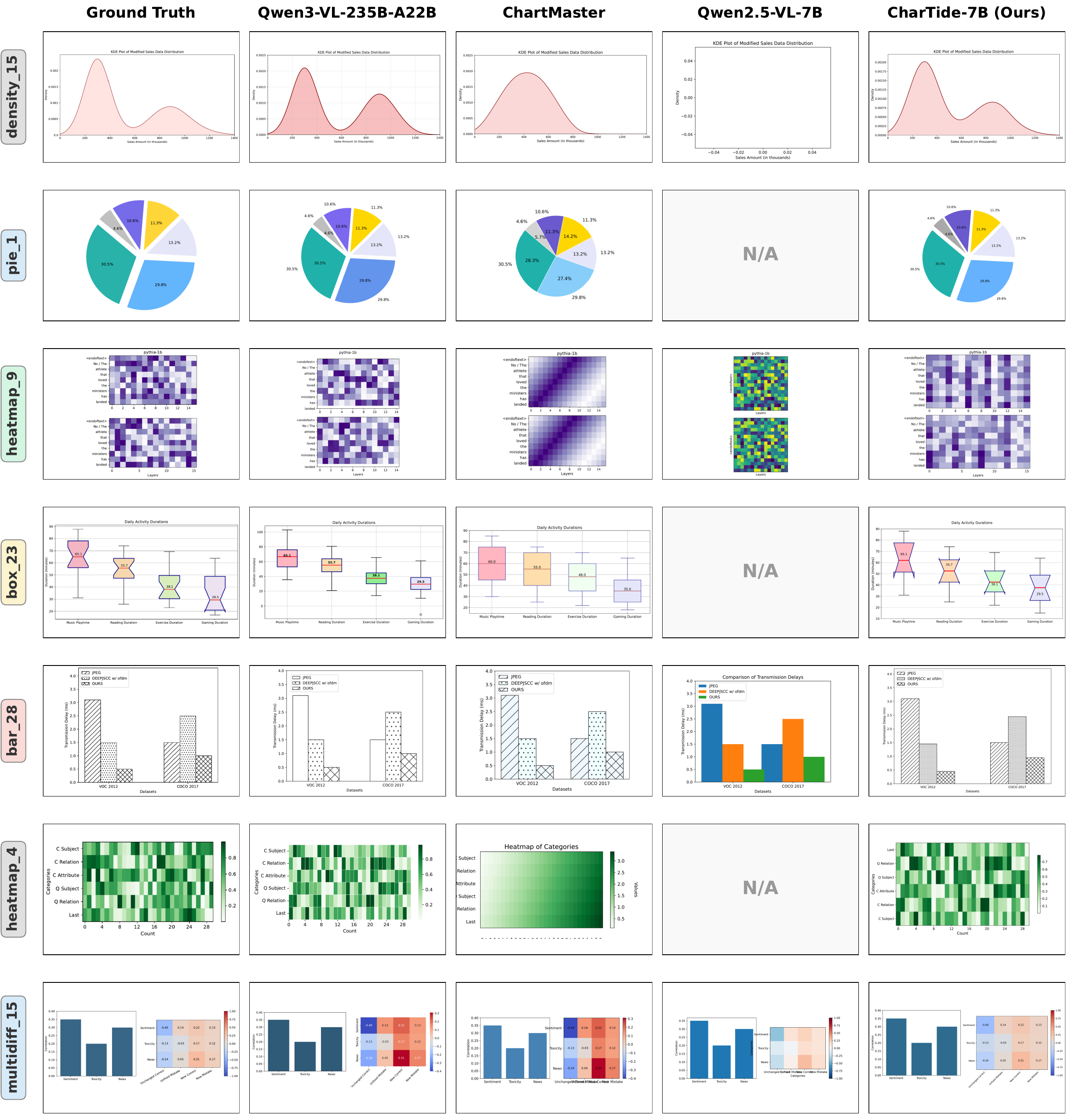}
    \caption{More visualization results on ChartMimic benchmark.  \textbf{"N/A"} in the figure indicates that the code generated by the model failed to compile, resulting in no output image. [1 / 3].}
    \label{fig:appendix_vis_1}
\end{figure}

\begin{figure}[h]
    \centering
    \includegraphics[width=1.0\textwidth]{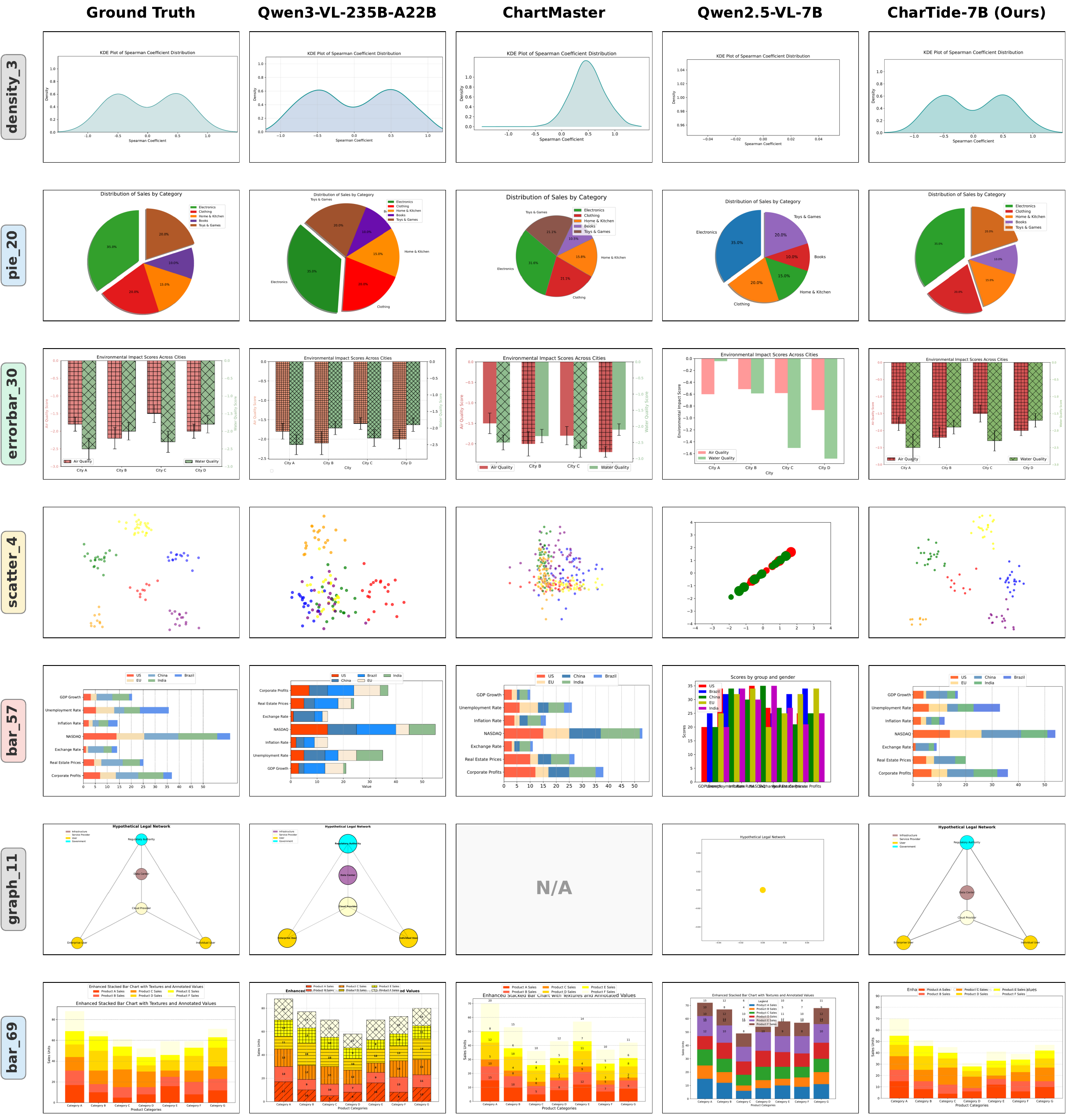}
    \caption{More visualization results on ChartMimic benchmark.  \textbf{"N/A"} in the figure indicates that the code generated by the model failed to compile, resulting in no output image. [2 / 3].}
    \label{fig:appendix_vis_2}
\end{figure}

\begin{figure}[h]
    \centering
    \includegraphics[width=1.0\textwidth]{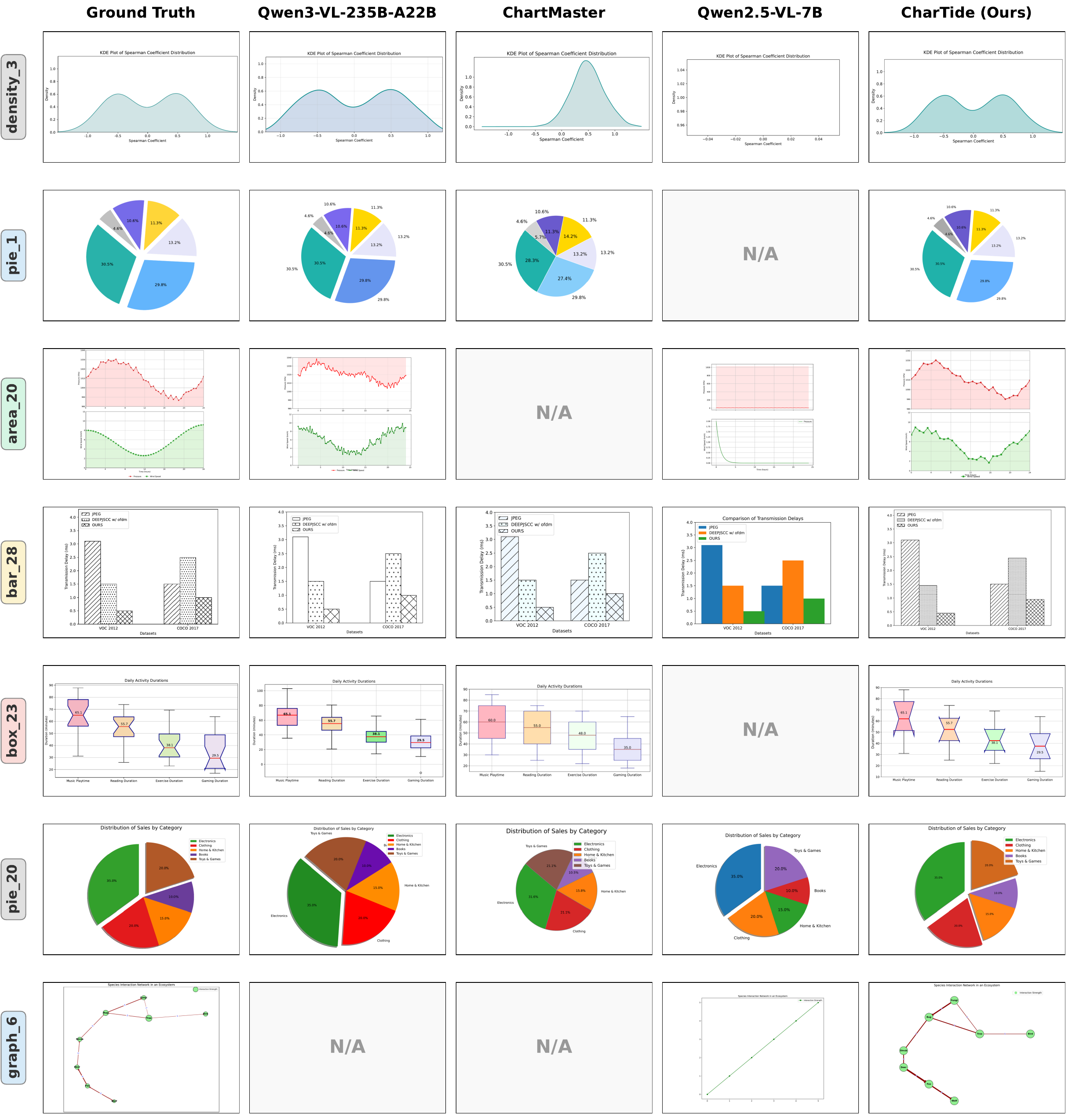}
    \caption{More visualization results on ChartMimic benchmark.  \textbf{"N/A"} in the figure indicates that the code generated by the model failed to compile, resulting in no output image. [3 / 3].}
    \label{fig:appendix_vis_3}
\end{figure}

\twocolumn

\end{document}